\documentclass[twoside,leqno,twocolumn]{article}

\usepackage[letterpaper]{geometry}

\usepackage{ltexpprt}
\usepackage{hyperref}
\usepackage{amsmath}
\usepackage[graphicx]{realboxes}
\usepackage{amssymb}
\usepackage{multirow}
\usepackage{makecell}
\usepackage{subfig}
\usepackage{xcolor}

\usepackage{algorithm}
\usepackage{algpseudocode}
\algblock{Input}{EndInput}
\algnotext{EndInput}
\algblock{Output}{EndOutput}
\algnotext{EndOutput}

\begin{document}

\title{\Large REGE: A Method for Incorporating Uncertainty in Graph Embeddings}
\author{Zohair Shafi\thanks{shafi.z@northeastern.edu} \\ \small{Northeastern University}\\ \small{Boston, MA, U.S.A.}
\and Germans Savcisens\thanks{g.savcisens@northeastern.edu}\\ \small{Northeastern University}\\ \small{Boston, MA, U.S.A.} \and Tina Eliassi-Rad\thanks{t.eliassirad@northeastern.edu} \\ \small{Northeastern University}\\ \small{Boston, MA, U.S.A.}}

\date{}

\maketitle







\begin{abstract}
Machine learning models for graphs in real-world applications are prone to two primary types of uncertainty: (1) those that arise from incomplete and noisy data and (2) those that arise from uncertainty of the model in its output. These sources of uncertainty are not mutually exclusive. Additionally, models are susceptible to targeted adversarial attacks, which exacerbate both of these uncertainties. In this work, we introduce \textbf{R}adius \textbf{E}nhanced \textbf{G}raph \textbf{E}mbeddings (REGE), an approach that measures and incorporates uncertainty in data to produce graph embeddings with radius values that represent the uncertainty of the model's output. REGE employs curriculum learning to incorporate data uncertainty and conformal learning to address the uncertainty in the model's output. In our experiments, we show that REGE's graph embeddings perform better under adversarial attacks by an average of 1.5\% (accuracy) against state-of-the-art methods.

\end{abstract}

\section{Introduction}

A graph embedding function, $f$, takes a graph $G=(V,E)$ with $n$ nodes, a node-attribute matrix $X \in \mathbb{R}^{n\times f}$, and the number of embedding dimensions $d$ and produces an embedding matrix $Z \in \mathbb{R}^{n \times d}$.\footnote{Graph neural networks (GNNs) also fall under the category of graph embedding functions. We use the terms graph embedding and node embedding interchangeably.} These embedding functions are useful in graph machine learning tasks such as node classification and link prediction and span various domains from financial systems \cite{wang2021review} to biological networks \cite{chatterjee2023improving,zhang2024hyganno} to social media analysis \cite{ying2018graph}.

A shortcoming of existing graph embedding functions is as follows: \textit{They do not capture the uncertainty that comes with incomplete and noisy graph data and/or the uncertainty that comes with the output of a model.} For example (in the latter case), can we guarantee that a given node $u$ in a graph $G$ maps exactly to $\vec{z}_u$ in a given $d$-dimensional space? We conjecture that the answer to this question is no. To address this, we propose a method that takes an existing graph embedding function and incorporates such uncertainties into the training stage, thereby generating an embedding $\vec{z}_u$ for each node $u$ as well as a radius value $r_u$ representing the uncertainty around the node's embedding. We further show that incorporating these uncertainties improves the accuracy of graph embeddings under adversarial attacks.

\paragraph{How do we define uncertainty in the graph data?} We use the low-rank representation of the graph $G$ to produce multiple instances of it: $\{G_1, G_2, G_3, \dots\}$. To do so, we find the eigen-decomposition of $G$. We then reconstruct a sequence of graphs using various low-rank representations of the original graph $G$. Next, we send the generated graph sequence, along with the original $G$, to a \textit{consensus} function that calculates the consensus in the presence/absence of edges in $G \,\cup\, \{G_1, G_2, G_3, \dots\}$. This procedure outputs a weighted adjacency matrix $W$, where each weight indicates the uncertainty associated with the presence of each edge. Finally, we pass $W$ to a \textit{radius} function that aggregates each row of $W$ and produces the uncertainty value for the node representing that row. The \textit{radius} function outputs a vector $\vec{r} \in \mathbb{R}^n$.

\paragraph{How do we define uncertainty in the model's output?} We utilize the student-teacher model \cite{hinton2015distilling, zhanggraph}. Specifically, we generate a teacher that takes the original graph $G$ and produces a graph embedding $Z$. We then train a multi-layer perceptron (MLP) using the inputs and outputs of the teacher. The trained MLP is called the \textit{student}. The student learns to predict $\alpha$-quantiles and $(1-\alpha)$-quantiles for each dimension of $\vec{z}_u$. (Recall that $\vec{z}_u$ is a row in $Z$ the output of the teacher function.)
Subsequently, we send the student's prediction to a conformal learning algorithm~\cite{angelopoulos2021gentle} that refines the predicted quantiles. The distance between the quantiles averaged across all dimensions is our definition of uncertainty in the graph embedding function's output.

\begin{figure*}
    \centering
    \includegraphics[width=1\linewidth]{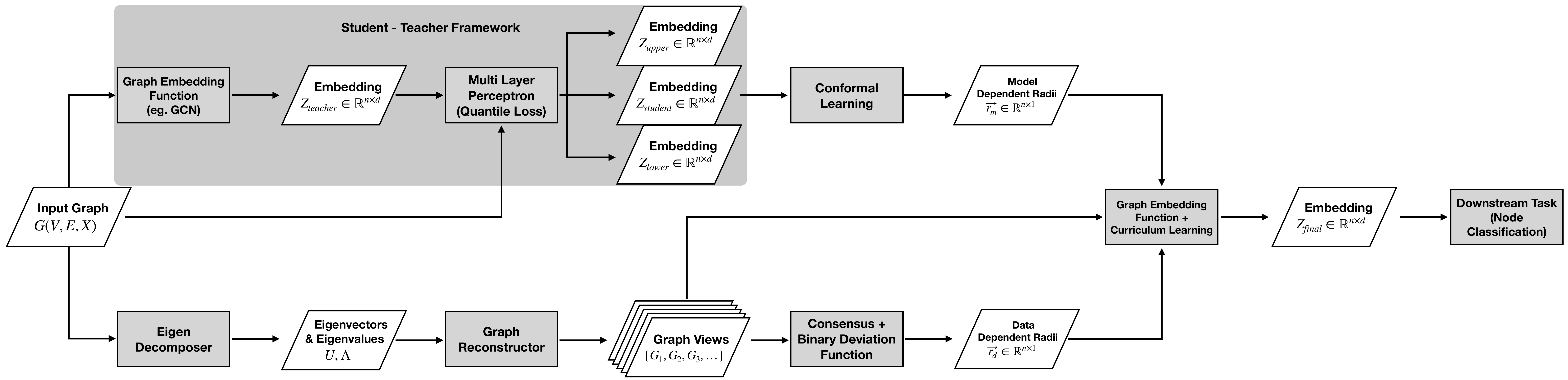}
    \caption{REGE uses eigen-decomposition to generate multiple graph views by systematically reconstructing the graph with an increasing number of components. These views are used to compute data-dependent radii (DDR) using a consensus and binary deviation function. A student-teacher model with quantile loss is used to capture uncertainty around each embedding dimension for each node to compute model-dependent radii (MDR). DDR or MDR is then incorporated into the training procedure as noise that is injected into the hidden layer representations. The generated graph views are used to train the graph embedding function using curriculum learning. Doing so leads to embeddings with higher accuracy under adversarial attacks on node classification.}
     \label{fig:pipeline}
\end{figure*}

\paragraph{How do we incorporate data uncertainty and the uncertainty in the model's output into the training procedure of a graph embedding function?} We inject noise into the hidden representation of each node proportional to the uncertainty values (a.k.a. ``radius'') for that node. We then use curriculum learning~\cite{bengio2009curriculum} on the graphs $G \cup \{G_1, G_2, G_3, \dots\}$. The curriculum learning improves the accuracy of the model under adversarial attacks. 

\paragraph{How do we evaluate?}
To evaluate our method, we compute the accuracy of node classification in the presence of attacks \cite{xu2019topology, zugner_adversarial_2019, liu2022towards} that perturb the structure of the input graph. The attacker wants to reduce the overall performance (measured by accuracy) of the graph embedding function. We show that our method outperforms multiple state-of-the-art methods. 


Our \textbf{contributions} are as follows:
\begin{itemize}
    \item We propose \textbf{R}adius \textbf{E}nhanced \textbf{G}raph \textbf{E}mbeddings (REGE), an approach that measures and incorporates data and model uncertainties when producing graph embeddings.
    \item REGE quantifies data-induced uncertainty using eigen-decompositions and model-induced uncertainty using a student-teacher model with conformal learning and curriculum learning. 
    \item Our experiments show that REGE outperforms current state-of-the-art methods in defending against adversarial attacks on node classification tasks by an average of 1.5\% (accuracy) across datasets and attacks.
   
\end{itemize}

\section{Proposed Method: REGE}

Given a graph $G=(V, E)$ with $|V| = n$ nodes, REGE measures and incorporates uncertainty in input data and model outputs by (1) computing a radius $r_i \in [0, 1]$ for each node $i \in V$ and (2) incorporating these radii into the training procedure. For a given node $i$, a radius value of $r_i = 0$ signifies minimum uncertainty, and $r_i = 1$ signifies maximum uncertainty. 

In REGE, radii are derived from two perspectives: Data-Dependent Radii (DDR), which capture uncertainty from the data itself, and Model-Dependent Radii (MDR), which reflect uncertainty in the model's output. The REGE pipeline is shown in Figure \ref{fig:pipeline}. In the following, we provide a detailed explanation of each step involved in the REGE approach.

\begin{figure}[h]
    \centering
    \includegraphics[width=1\linewidth]{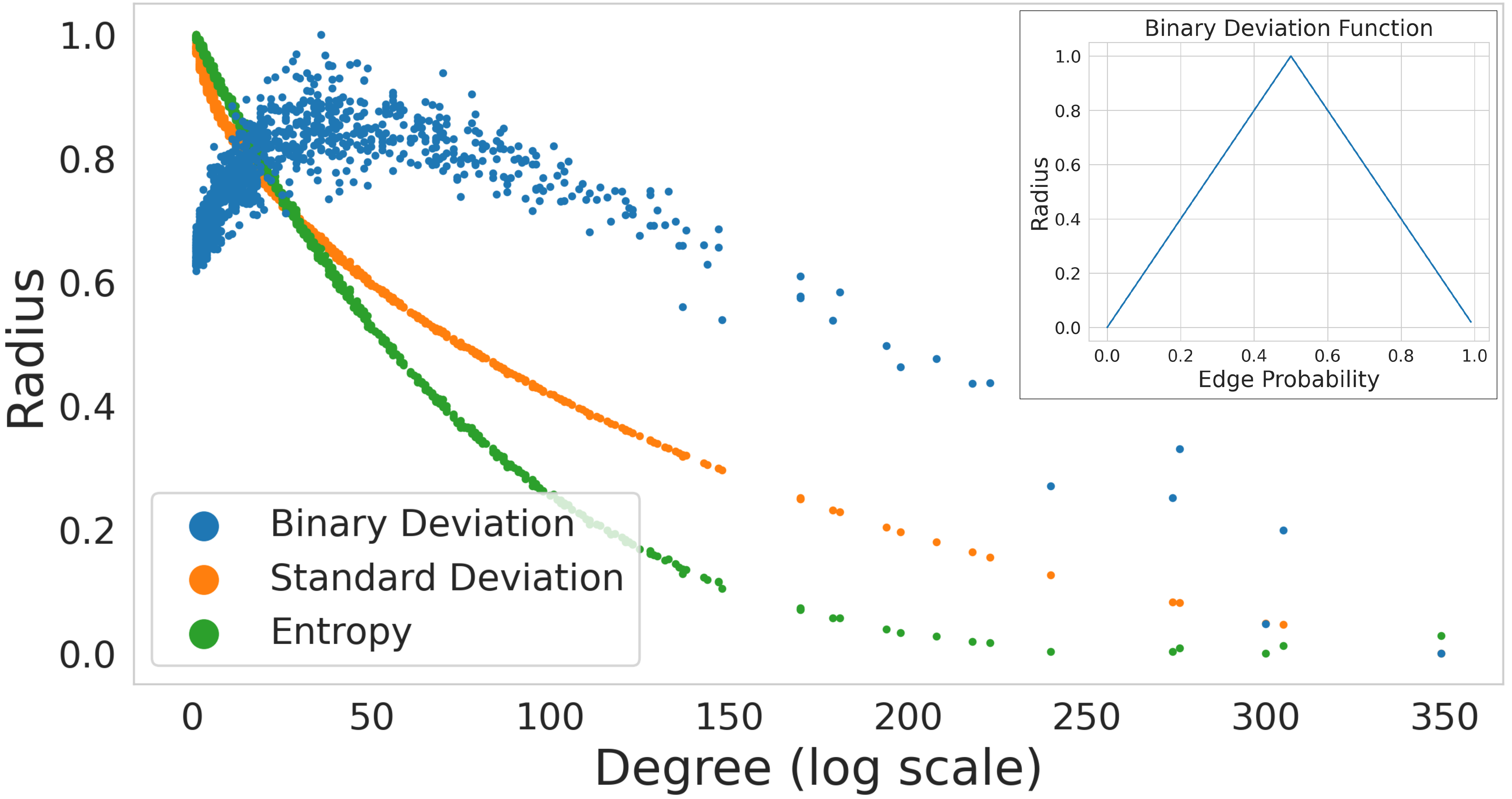}
   
    \caption{Radii for each node in the PolBlogs network, derived using binary deviation, standard deviation, and entropy functions, plotted against node degree. Unlike standard deviation and entropy, which correlate radii closely with degree, the binary deviation function allows low-degree nodes to also have low radii, capturing more information than just degree distribution. (inset) The binary deviation function calculates uncertainties based on entries of the weighted adjacency matrix $W$. For an edge between nodes $i$ and $j$, $W_{ij} = 0$ indicates that the edge was never observed, while $W_{ij} = 1$ indicates the edge was always observed, both representing high certainty. In contrast, $W_{ij} = 0.5$ corresponds to the highest level of uncertainty.}
     \label{fig:binary_deviation}
\end{figure}

\subsection{Data-Dependent Radii}



Uncertainty in graph data can stem from (1) uncertainty in the labels, (2) uncertainty in the node feature matrix $X$, and/or (3) uncertainty in the edges (structural uncertainty). \footnote{We do not consider uncertainty in higher-order structures such as triangles or motifs.} In this paper, we address uncertainty in the edges -- i.e., uncertainty arising fr In this paper, we address uncertainty in the edges -- i.e., the uncertainty that arises from noisy interactions between the nodes. This noise can have various causes, e.g., missing values, data acquisition problems, or observation errors. We make no assumptions about the nature and distribution of this noise. In this context, our main focus is \textbf{not} on the generative processes that produce the graph, but on effectively mitigating the effects of noisy edges in the training stage. We capture the uncertainty of the edges and propagate these edge uncertainties to the incident nodes. Furthermore, we evaluate REGE by testing it against adversarially designed noise (instead of random noise), which is introduced specifically to reduce the overall classification accuracy of the model.

Dealing with uncertainty on edges depends on whether one has access to the underlying data from which a network is derived or whether a fixed network is given. The problem of inferring networks from noisy data is well-studied. For example, Young et al.~\cite{young2020bayesian} propose a general framework for inferring network structure from noisy measurements using Bayesian inference, which provides a posterior distribution over possible network structures. Peixoto et al.~\cite{peixoto2024network} also use Bayesian inference with minimum description length \cite{barron1998minimum} to reconstruct networks from data without requiring prior knowledge of the weight distribution or the number of edges.

We focus on the case where a fixed network structure is given, as is the case for many publicly available graph datasets. We use the eigen decomposition of the fixed network structure to obtain an estimate of the uncertainty on the edges. We build on the insight that reconstructing a graph from eigenvectors associated with large eigenvalues reconstructs a graph that has edges with high certainty \cite{sawlani2021fast, cucuringu2011localization, mitrovic2009spectral}.


Based on this observation, REGE generates multiple views of the input graph $G$ by reconstructing it from the eigen-decomposition of its adjacency matrix $A$. The variance, or ``energy'', of a reconstructed graph can be defined as : 
\begin{align}
    E &= \frac{\sum_{i = 1}^q \lambda_i^2}{\sum_{i = 1}^n \lambda_i^2}
\end{align}
where $\lambda_i$ is the $i-$th eigenvalue, $q$ is the number of components used to reconstruct the graph, and $n$ is the total number of components. Concretely, if $q=10$, then $G_1$ is reconstructed using the first 10 components. As $q$ increases, more energy is retained. REGE takes as input a minimum value of $q$ to reconstruct the first graph $G_1$, then systematically adds more components to reconstruct graphs $G_2, G_3, \dots$, until all components are included. This results in a series of reconstructed graphs $\{G_1, G_2, G_3,\dots\}$ that capture different levels of structural detail.\footnote{If the underlying data (instead of the network structure) is available, the graph views $\{G_1, G_2, \dots\}$ can be sampled using the method described by Young et al.~\cite{young2020bayesian}.}

A consensus among these generated graphs is constructed as a weighted adjacency matrix $W$ by averaging the adjacency matrices of the reconstructed graphs. Specifically, for each edge, the corresponding matrix entry $W_{ij}$ represents the frequency with which that edge appears across all reconstructed views. A value close to $1$ indicates that the edge is highly certain and appears consistently in all views, while a value close to $0$ indicates that the edge is highly unlikely. On the other hand, a value of $0.5$ reflects the highest level of uncertainty, as the edge is observed only in half of the views. This edge-level uncertainty directly impacts the node-level uncertainty. Nodes with many uncertain edges in their local neighborhood (i.e., the row of $W$ that corresponds to the given node) are likely to have increased uncertainty in their representations. 

Formally, let $e$ be an edge between nodes $i$ and $j$, then $W_{ij}$ denotes the entry in $W$ for edge $e$. The uncertainty $u_e$ for the edge is computed using a binary deviation function, defined as: 
\begin{align}
    u_e = 1 - \Big|W_{ij} - (1 - W_{ij})\Big|
\end{align}
This function captures the maximum uncertainty when $W_{ij}$ is $0.5$. The radius $r_i$ for a node $i$ is then calculated by averaging the uncertainty of the edges incident on the node. This captures the overall data uncertainty surrounding each node.

Figure \ref{fig:binary_deviation} provides an intuitive visualization of radii by plotting them against node degrees for the PolBlogs network \cite{adamic2005political}. Here, we also compute radii using the row-wise standard deviation and entropy of the weighted adjacency matrix $W$. Observe that these functions tend to follow the degree distribution, whereas the binary deviation function allows low-degree nodes to also have low radii, thereby capturing more than just the degree distribution. 

\begin{algorithm}[h]
\caption{Generate Radius Values from eigen-decomposition}
\label{radii_alg}
\begin{algorithmic}[1]

    \Require Adjacency matrix $A$ of graph $G$ with $n$ nodes, minimum number of components $q$ to reconstruct the first graph 
    
    \State \textbf{Step 1: Compute Eigen Decomposition}
    \State $U, \Lambda, V \gets \text{eig}\,(A)$ \Comment{Compute eigenvalues and eigenvectors of $A$}
    
    \State \textbf{Step 2: Regenerate $c$ views of the graph with increasing number of components. The maximum number of components is n.}
    \For{$k \in \{q, q+1, \ldots, n\}$}
        \State $A_k \gets U[:, 1:k] \cdot \Lambda[1:k, 1:k] \cdot V[1:k, :]$
        \State Normalize $A_k$ to $[0, 1]$
        \State Append $A_k$ to the list of sampled graphs
    \EndFor
    
    \State \textbf{Step 3: Create weighted adjacency matrix $W$}
    \State Initialize an empty matrix $W$ of the same size as $A$
    \For{each pair of nodes $(i, j)$}
        \State $W[i, j] \gets \frac{1}{c} \sum_{k=1}^c (A_k[i, j]) \,\forall\, j \neq i$
    \EndFor
    
    \State \textbf{Step 4: Compute radius values from the generated graph}
    
    \State Initialize an empty list $\vec{r}$ for radius values
    \For{each node $i$}
        \State $r_i = 1 - \frac{1}{n} \sum |W_{i,:} - (1 - W_{i, :})|$
        \Comment{Binary Deviation}
        \State Append $r_i$ to $\vec{r}$
    \EndFor
    
    \State \textbf{Return} $\vec{r}$ \Comment{Vector of radius values for each node}
\end{algorithmic}
\end{algorithm}

Formally, given $c$ views of the graph, with corresponding adjacency matrices $A_1, A_2, \dots, A_c$, the radii around each node is calculated as : 

\begin{align}
    W &= \frac{1}{c}\sum_{k = 1}^c A_k \\
    r_i &= 1 - \frac{1}{n} \sum \Big|W_{i,:} - (1 - W_{i, :})\Big|
\end{align}

where, $W$ is the weighted adjacency matrix, and $r_i$ is the radius value of node $i$.

Algorithm \ref{radii_alg} contains the algorithm for generating data-dependent radii.


\subsection{Model-Dependent Radii}
Uncertainty in the model's output, which can arise from the parameters, optimizing strategy, the upper limit of the model fitting ability, or the lack of training knowledge \cite{zhou2022survey}, is a critical factor in assessing the reliability of predictions \cite{hullermeier2021aleatoric}. Rather than measuring uncertainty at the edge level and propagating it to the node level as previously done, we measure uncertainty in the model's output across each dimension of a node's embedding space. Specifically, for a node represented as a $d$-dimensional embedding vector, we calculate a ``radius'' of uncertainty for \textit{each} dimension.

To achieve this, REGE uses conformal learning \cite{angelopoulos2021gentle}, a method that provides statistically rigorous prediction intervals. Conformal learning uses a calibration set of data points (different from validation or testing data) to create prediction sets for any model. The calibration process determines thresholds for the prediction intervals, ensuring that the intervals meet the desired coverage with statistical guarantees on unseen data.

We focus on conformal quantile regression, as described by \cite{angelopoulos2021gentle}. Quantile regression is a technique used to estimate conditional quantiles of a continuous target variable $y \in \mathbb{R}^1$ given an input vector $x \in \mathbb{R}^{d}$. Specifically, quantile regression aims to estimate the $\gamma$-th percentile of the target variable, denoted by $f_{\gamma}(x)$, which represents the value below which a fraction $\gamma$ of the distribution of $y$ falls.

For example, $f_{.05}(x)$ corresponds to the 5th percentile of $y$ given $x$ and, $f_{.95}(x)$ corresponds to the 95th percentile. Thus, the interval $[f_{.05(x)}, f_{.95(x)}]$ should capture 90\% of the possible values of $y$. However, in practice, the learned quantile function $\hat{f}_\gamma(x)$ may deviate from the true quantiles, leading to intervals that either under or overestimate the actual coverage. 

Conformal quantile regression addresses this issue by adjusting the intervals to ensure that they meet the desired coverage level. This is achieved by \textit{conformalizing} the intervals, i.e., shrinking or expanding their range based on a calibration set, thus guaranteeing that the intervals cover the true variable $y$ with the specified probability. 

Formally, given a desired coverage ratio $\alpha$ and two learned regression models for the upper and lower quantiles $\hat{f}_{1-\frac{\alpha}{2}}(x)$ and $\hat{f}_{\frac{\alpha}{2}}(x)$, we define the conformal score $s$ as:
\begin{align}
    s(x, y) = max\Big\{\hat{f}_{\frac{\alpha}{2}}(x) - y, y - \hat{f}_{1 - \frac{\alpha}{2}}(x)\Big\}
\end{align}

Conformal scores are calculated for each of the $n$ calibration data points to produce scores $s_1, s_2, \dots, s_n$. Finally, prediction intervals are updated as:
\begin{align}
\label {qhat}
    C(x) = \Big[\hat{f}_{\frac{\alpha}{2}}(x) - \hat{q}, \hat{f}_{1- \frac{\alpha}{2}}(x) + \hat{q}\Big]
\end{align}
where 
\begin{align}
    \hat{q} = Quantile\,\Big(s_1, s_2, \dots, s_n; \frac{\lceil (n + 1)(1 - \alpha) \rceil}{n}\Big)
\end{align}
Notice that conformalized quantile regression shrinks or grows the interval by $\hat{q}$ to achieve guaranteed coverage.

In the context of REGE, conformal quantile regression is applied to estimate uncertainty at each dimension of the embedding for every node in the graph. For a node embedded in a $d$-dimensional space, we generate an uncertainty value (or radius) for each dimension. This process is facilitated by a student-teacher model  \cite{hinton2015distilling, zhanggraph}. Initially, a Graph Convolutional Network (GCN) (or any other graph embedding function) is trained on the node classification task. Following this, a simple multi-layer perceptron (MLP) (the student) is trained to predict the outputs of the GCN trained earlier (the teacher), as well as to estimate the quantile ranges for each dimension. The student is trained using the quantile loss given by:
\begin{align}
    \mathcal{L}(y, \hat{y}, q) = max\Big\{(q - 1) \cdot (y - \hat{y}), q \cdot (y - \hat{y})\Big\}
\end{align}
where $q$ is the desired quantile (set to $\frac{\alpha}{2}$ or $1 - \frac{\alpha}{2}$), $y$ the true value, and $\hat{y}$ the predicted value. We set $\alpha = 0.05$. 

Conformal learning is then used to refine these intervals (using the training set as the calibration set) as shown in Equation \ref{qhat}. The final radius at the node level is calculated by averaging the dimensional uncertainties for each node.


\begin{algorithm}
\caption{Training GCN with Curriculum Learning and Radius Noise}
\label{alg_train}
\begin{algorithmic}[1]
\Require Adjacency matrix $A$ of graph $G$ with $n$ nodes, Feature matrix $X_{input}$, Radius values $\vec{r}$, number of epochs

\State \textbf{Initialize} GCN model parameters

\For{$k \in \{q, q+1, \ldots, n\}$} \Comment{Curriculum Learning with increasing components}
    \For{each epoch}
        \State $X \gets X_{input}$
        \For{each GCN Layer}
            \State $X \gets \text{GCN}(X, A_k)$ \Comment{GCN trained on $A_k$ instead of $A$}
            \State $X \gets X + \mathcal{N}(0, \vec{r})$ \Comment{Add noise sampled from normal distribution with variance set to radius values}
        \EndFor
        \State \textbf{Compute loss}
        \State $loss \gets \text{Loss}(X, \text{labels})$
        \State \textbf{Backpropagate and Update Parameters}
    \EndFor
\EndFor

\State \textbf{Return} Trained GCN model
\end{algorithmic}
\end{algorithm}

\subsection{Incorporating Radii}
We incorporate the generated radii (either data-dependent radii or model-dependent radii) by perturbing the node embeddings based on the radii associated with each node. Specifically, we perturb the hidden layer representations of each node by adding noise proportional to the node’s radius, thereby reflecting the uncertainty associated with that node in the graph.

The noise is sampled from a normal distribution with a mean of zero and a variance equal to the node's radius. This approach ensures that nodes with low variance (i.e., lower uncertainty) maintain relatively stable embeddings, while nodes with higher variances (i.e., higher uncertainty) exhibit less stable embeddings. This controlled instability compels the model to learn robust representations for uncertain nodes. This is analogous to variational autoencoders (VAEs) \cite{kingma2013auto}, where the representations are sampled from a distribution with means and variances learned through backpropagation. However, unlike VAEs, in REGE, the variances (or radii) are fixed. Formally, for each node $i$, radius $r_i$ and hidden layer representation $x^l_i$ at layer $l$, we update the representation as:
\begin{align}
    x^l_i \gets x^l_i + \mathcal{N}(0, r_i)
\end{align}

\paragraph{Training Method}
We outline the final training procedure for REGE, which uses curriculum learning \cite{bengio2009curriculum}. Recall that the graphs reconstructed with the fewest components are simpler, containing fewer edges that are associated with the highest certainty. In the curriculum training process, we train the model sequentially on these graph views, starting with the most simplified version (e.g., the graph reconstructed using only $10$ components). We then progressively introduce more complexity by training on the graph views reconstructed with an increasing number of components (e.g., $20$ components, and so on) until the full graph is utilized or early stopping is triggered based on the patience criteria. This gradual introduction of complexity helps the model learn robust representations at each stage. The complete details of this training procedure are provided in Algorithm \ref{alg_train}.

\section{Experiments}
\subsection{Experimental Setup}

We evaluate REGE using three widely-used datasets: Cora \cite{mccallum2000automating}, Citeseer \cite{giles1998citeseer}, and PolBlogs \cite{adamic2005political}. To assess REGE's robustness, we measure its performance when the graph structure is attacked. Concretely, we use three well-known non-targeted availability attacks on graph structure: MinMax \cite{xu2019topology}, Meta-Attack \cite{zugner_adversarial_2019} and GraD \cite{liu2022towards} - with perturbations of 1\% and 10\%, reflecting realistic attack scenarios (we provide results for additional perturbation ranges and the CoraML \cite{mccallum2000automating} dataset in the Appendix). We use the CE-PGD variant MinMax and adversarial graphs for Meta-Attack were computed using a standard GCN surrogate model. 

We compare REGE against eight defense methods: RGCN \cite{zhu2019robust}, Adversarial Training (Adv-Train) \cite{xu2019topology}, GCN-SVD \cite{entezari2020all}, ProGNN \cite{jin2020graph}, GNNGuard \cite{zhang2020gnnguard}, GraphReshape \cite{wang2024graph}, GADC \cite{liugraph}, and Ricci-GNN \cite{yericci}. This ensures coverage across a wide range of defense strategies.
\begin{itemize}
    \item  RGCN \cite{zhu2019robust} and Adversarial Training \cite{xu2019topology} modify the learning method. RGCN uses Gaussian distributions as the hidden layer representation in each convolutional layer to improve robustness against attacks and adversarial training uses a bi-level objective function where the outer optimization trains the model to minimize its loss, while the inner optimization generates adversarial perturbations to maximize this loss. 
    \item GCN-SVD \cite{entezari2020all} and ProGNN \cite{jin2020graph} exploit low-rank properties of the graph. GCN-SVD uses a low-rank reconstruction of the input graph. ProGNN learns a clean graph by reconstructing it with sparsity and low-rank constraints in conjunction with the downstream task objective.
    \item GNNGuard \cite{zhang2020gnnguard} and GraphReshape \cite{wang2024graph} examine edges in the graph. GNNGuard assigns higher weights to edges connecting similar nodes while pruning edges between dissimilar nodes. GraphReshape changes edges around each node to shift the classification boundary of the learned model.
    \item GADC \cite{liugraph} and Ricci-GNN \cite{yericci} reconstruct the input graphs.  GADC uses adversarial training and Graph Diffusion Convolution \cite{gasteiger2019diffusion}. Ricci-GNN uses the Ricci flow metric \cite{ollivier2009ricci} to generate multiple views of a graph and train on each view.
\end{itemize}

Our experiments include two variants of REGE — one utilizing data-dependent radii (REGE (D)) and the other employing model-dependent radii (REGE (M)).

We use GCNs as our graph embedding functions. The model architecture consists of two GCN layers with 16 hidden units, followed by a dropout layer (with $p = 0.5$) after the first GCN layer. The teacher model uses the same architecture, while the student model is an MLP with three hidden layers, each containing $1024$ neurons, with dropout layers (with $p = 0.5$) after the first two dense layers. We make our implementation available at \url{https://github.com/zohairshafi/REGE}. 





\subsection{Results and Discussion}
Tables \ref{tab:citeseer_tab}, Table \ref{tab:cora_tab}, and Table \ref{tab:polblogs_tab} summarize the results for the Citeseer, Cora, and PolBlogs datasets, respectively. They show the mean node classification accuracy over 10 runs, along with standard deviations.  REGE consistently outperforms competing methods in terms of accuracy by an average of 1.5\%. We show more results in the Appendix. Those results are similar to the ones shown here.

\begin{table*}[h]
    \centering
    \begin{tabular}{c|cc|cc|cc}
        Method          & MinMax (1\%)    &MinMax (10\%)     & Meta (1\%)      & Meta (10\%)      & GraD (1\%) & GraD (10\%) \\
        \hline
GCN             & 	$.691\pm.004$ 	&	$.561\pm.004$  &	 $.717\pm.004$ 	&	$.629\pm.003$  & $.678\pm.004$ 	&	$.646\pm.006$  \\
Adv-Train             & 	$.659\pm.001$ 	&	$.660\pm.001$  &	 $.659\pm.002$ 	&	$.658\pm.001$  & $.577\pm.001$ 	&	$.579\pm.001$  \\
RGCN            & 	$.714\pm.011$ 	&	$.633\pm.025$  &	 $.729\pm.006$ 	&	$.684\pm.008$  & $.689\pm.006$ 	&	$.664\pm.004$  \\
GCN-SVD         & 	$.673\pm.006$ 	&	$.652\pm.008$  &	 $.681\pm.007$ 	&	$.687\pm.008$  & $.644\pm.012$ 	&	$.660\pm.007$   \\
GNNGuard        & 	$.714\pm.007$ 	&	$.676\pm.008$  &	 $.721\pm.009$ 	&	$.666\pm.022$  & $.702\pm.005$ 	&	$.685\pm.007$  \\
ProGNN          & 	$.718\pm.007$ 	&	$.707\pm.008$  &	 $.723\pm.005$ 	&	$.700\pm.006$    & $.684\pm.007$ 	&	$.680\pm.007$   \\
GADC            & 	$.730\pm.008$ 		&	$.722\pm.005$  &	 $.737\pm.008$ 	&	$.727\pm.007$  & $.701\pm.003$ 	&	$\textbf{.701}\pm\textbf{.004}$  \\
GraphReshape    & 	$.717\pm.001$ 	&	$.712\pm.003$  &	 $.714\pm.005$ 	&	$.708\pm.005$  & $.693\pm.002$ 	&	$.695\pm.001$  \\
Ricci-GNN       & 	$.705\pm.007$ 	&	$.696\pm.006$  &	 $.724\pm.004$ 	&	$.716\pm.008$  & $.667\pm.005$ 	&	$.666\pm.005$  \\
\hline
REGE (D)        & 	$\textbf{.749}\pm\textbf{.002}$ 	&	$\underline{.735\pm.003}$  &	 $\textbf{.751}\pm\textbf{.003}$ 	&	$\underline{.75\pm.003}$   & $\textbf{.72}\pm\textbf{.004}$ 	&	$\underline{.699\pm.011}$   \\
REGE (M)        & 	$\underline{.736\pm.014}$ 	&	$\textbf{.74}\pm\textbf{.002}$   &	 $\underline{.738\pm.037}$ 	&	$\textbf{.752}\pm\textbf{.002}$  & $\underline{.715\pm.003}$ 	&	$.696\pm.057$  \\
    \end{tabular}
    \caption{Node classification accuracy on Citeseer averaged over 10 runs, with standard deviations. REGE-D (data-dependent radii) and REGE-M (model-dependent radii) consistently outperform competing methods. Best results are in \textbf{bold}, with the second-best \underline{underlined}. Comparisons include older methods such as RGCN and GCN-SVD, and newer methods such as GADC and GraphReshape.}
    \label{tab:citeseer_tab}
\end{table*}

\begin{table*}[h]
    \centering
    \begin{tabular}{c|cc|cc|cc}
        Method          & MinMax (1\%)    &MinMax(10\%)     & Meta (1\%)      & Meta(10\%)      & GraD (1\%) & GraD (10\%) \\
        \hline
       GCN          &    $.799\pm.002$    &       $.678\pm.004$     &        $.806\pm.002$        &       $.618\pm.006$         &        $.794\pm.005$    &       $.793\pm.004$ \\
       Adv-Train             & 	$.755\pm.002$ 	&	$.757\pm.001$  &	 $.758\pm.001$ 	&	$.763\pm.002$  & $.759\pm.001$ 	&	$.758\pm.001$  \\

RGCN         &   $.824\pm.005$    &       $.763\pm.007$     &        $.824\pm.004$        &       $.694\pm.006$         &        $.829\pm.006$    &       $.803\pm.005$ \\
GCN-SVD      &   $.720\pm.005$     &       $.708\pm.006$     &        $.716\pm.003$        &       $.700\pm.005$           &        $.712\pm.005$    &       $.688\pm.008$ \\
GNNGuard     &   $.825\pm.007$    &       $.785\pm.009$     &        $.816\pm.007$        &       $.705\pm.016$         &        $.834\pm.005$    &       $.808\pm.004$ \\
ProGNN       &   $.831\pm.003$    &       $\textbf{.830}\pm\textbf{.004}$        &        $.830\pm.004$         &       $\underline{.801\pm.006}$         &        $.807\pm.006$    &       $.805\pm.004$ \\
GADC         &   $.790\pm.004$     &       $.784\pm.004$     &        $.787\pm.005$        &       $.779\pm.004$         &        $.791\pm.004$    &       $.785\pm.006$ \\
GraphReshape &   $.834\pm.003$    &       $.800\pm.005$       &        $.835\pm.000$          &       $.788\pm.000$           &        $.829\pm.006$    &       $.798\pm.003$ \\
Ricci-GNN    &   $.835\pm.005$    &       $.797\pm.006$     &        $.832\pm.004$        &       $\underline{.801\pm.004}$         &        $.829\pm.004$    &       $.793\pm.004$ \\
\hline
REGE (D)     &   $\underline{.837\pm.004}$    &       $.806\pm.006$     &        $\underline{.841\pm.003}$        &       $\textbf{.835}\pm\textbf{.003}$       &        $\underline{.842\pm.005}$    &       $\underline{.809\pm.004}$ \\
REGE (M)     &   $\textbf{.838}\pm\textbf{.003}$  &       $\underline{.808\pm.005}$     &        $\textbf{.843}\pm\textbf{.004}$      &       $\textbf{.835}\pm\textbf{.004}$       &        $\textbf{.844}\pm\textbf{.004}$  &       $\textbf{.815}\pm\textbf{.003}$ \\
    \end{tabular}
    \caption{Node classification accuracy on Cora averaged over 10 runs, with standard deviations. REGE-D (data-dependent radii) and REGE-M (model-dependent radii) consistently outperform competing methods. Best results are in \textbf{bold}, with the second-best \underline{underlined}. Comparisons include older methods such as RGCN and GCN-SVD, and newer methods such as GADC and GraphReshape.}
    \label{tab:cora_tab}
\end{table*}

\begin{table*}[h]
    \centering
    \begin{tabular}{c|cc|cc|cc}
        Method          & MinMax (1\%)    &MinMax(10\%)     & Meta (1\%)      & Meta(10\%)      & GraD (1\%) & GraD (10\%) \\
        \hline
GCN             &    $.944\pm.001$     &  $.871\pm.002$     &    $.859\pm.002$    &    $.726\pm.004$    &    $.876\pm.005$    &   $.795\pm.002$ \\ 
Adv-Train             & 	$.909\pm.001$ 	&	$.862\pm.001$  &	 $.894\pm.003$ 	&	$.876\pm.001$  & $.820\pm.002$ 	&	$.806\pm.002$  \\
RGCN            &    $.936\pm.002$     &  $.854\pm.002$     &    $.850\pm.002$     &    $.699\pm.007$    &    $.866\pm.003$    &   $.811\pm.003$ \\ 
GCN-SVD         &    $.939\pm.005$     & \underline{$.885\pm.002$}  &    $.926\pm.002$    &    $.894\pm.007$    &    $.883\pm.004$    &   $\textbf{.865}\pm\textbf{.003}$ \\ 
GNNGuard        &    $\textbf{.950}\pm\textbf{.004}$        &  $.861\pm.001$     &    $.854\pm.002$    &    $.707\pm.014$    &    $.855\pm.005$    &   $.812\pm.002$ \\ 
ProGNN          &    $.935\pm.017$     &  $.869\pm.029$     &    \underline{$.936\pm.023$}    &    $.823\pm.055$    &    $.829\pm.029$    &   $.859\pm.005$ \\ 
GADC            &    $.512\pm.008$     &  $.512\pm.008$     &    $.512\pm.008$    &    $.512\pm.008$    &    $.498\pm.009$    &   $.497\pm.014$ \\ 
GraphReshape    &    $.935\pm.007$     &  $.847\pm.002$     &    $.850\pm.006$     &    $.694\pm.002$    &    $.851\pm.003$    &   $.803\pm.004$ \\ 
Ricci-GNN       &    $.941\pm.004$     &  $.874\pm.004$     &    $.932\pm.003$    &    $.928\pm.010$     &    $.875\pm.011$    &   $\textbf{.865}\pm\textbf{.008}$ \\ 
\hline
REGE (D)        &    \underline{$.946\pm.004$}     &  $\textbf{.890}\pm\textbf{.004}$        &    $\textbf{.946}\pm\textbf{.007}$      &    $\textbf{.950}\pm\textbf{.005}$   &    \underline{$.887\pm.002$}    &   $\textbf{.865}\pm\textbf{.003}$ \\ 
REGE (M)        &    $.929\pm.009$     &  $.880\pm.006$      &    $.931\pm.017$    &    \underline{$.942\pm.017$}    &    $\textbf{.889}\pm\textbf{.002}$  &   \underline{$.861\pm.004$} \\ 

    \end{tabular}
    \caption{Node classification accuracy on PolBlogs averaged over 10 runs, with standard deviations. REGE-D (data-dependent radii) and REGE-M (model-dependent radii) consistently outperform competing methods. Best results are in \textbf{bold}, with the second-best \underline{underlined}. Comparisons include older methods such as RGCN and GCN-SVD, and newer methods such as GADC and GraphReshape.}
    \label{tab:polblogs_tab}
\end{table*}

\subsection{Ablation Study}
Training REGE involves two components: integrating radii as noise and employing curriculum learning to sequentially train on graph views. Table \ref{tab:ablation_table} presents the results of an ablation study, averaged over 10 runs, where we systematically remove these components to assess their impact. The variant REGE-NCT (\textbf{N}o \textbf{C}urriculum \textbf{T}raining), which excludes curriculum training, still demonstrates competitive performance, surpassing several state-of-the-art methods. However, REGE outperforms REGE-NCT in most cases, highlighting the effectiveness of curriculum learning. For conciseness, the results presented are for 10\% perturbation levels. Results for CoraML and PolBlogs networks are shown in the Appendix (Table \ref{tab:ablation_table_2}). 

\begin{table*}[h]
    \centering
    \begin{tabular}{c|ccc|ccc}
             &                  &   Citeseer        &                   &                    &   Cora             &                 \\
Method       &   MinMax         &   Meta-Attack     &   GraD            &   MinMax           &   Meta-Attack      &   GraD          \\
\hline
REGE-NCT (D) & $.686\pm.011$    &   $.690\pm.007$    &   $\textbf{.702}\pm\textbf{.003}$   &   $.797\pm.009$    &    $.694\pm.011$   &  $.807\pm.003$  \\
REGE (D)     & $\textbf{.735}\pm\textbf{.003}$    &   $\textbf{.750}\pm\textbf{.003}$    &   $.699\pm.011$   &   $\textbf{.806}\pm\textbf{.006}$    &    $\textbf{.835}\pm\textbf{.003}$   &  $\textbf{.809}\pm\textbf{.004}$  \\
\hline
REGE-NCT (M) & $.704\pm.009$    &   $.694\pm.004$   &   $\textbf{.704}\pm\textbf{.003}$   &   $.804\pm.005$    &    $.699\pm.008$   &  $.805\pm.010$   \\
REGE (M)     & $\textbf{.740}\pm\textbf{.002}$     &   $\textbf{.752}\pm\textbf{.002}$   &   $.696\pm.057$   &   $\textbf{.808}\pm\textbf{.005}$    &    $\textbf{.835}\pm\textbf{.004}$   &   $\textbf{.815}\pm\textbf{.003}$ \\
    \end{tabular}
    \caption{Node classification results from the ablation study for Citeseer and Cora networks. The REGE-NCT variant, which excludes curriculum learning,  demonstrates superior performance compared to other state-of-the-art methods as shown in Tables \ref{tab:citeseer_tab} and \ref{tab:cora_tab}. However, REGE outperforms REGE-NCT in most cases, highlighting the effectiveness of incorporating both radii and curriculum learning into the model. Results presented are for 10\% perturbation levels averaged across 10 runs. Best results are in \textbf{bold}. Results for CoraML and PolBlogs are consistent with the results shown above and are shown in the Appendix in Table \ref{tab:ablation_table_2}.}
    \label{tab:ablation_table}
\end{table*}

\subsection{Comparing Data and Model-Dependent Radii}
\begin{figure}[h!]
    \centering
    \subfloat[Karate Club Network]{\includegraphics[width=0.72\linewidth]{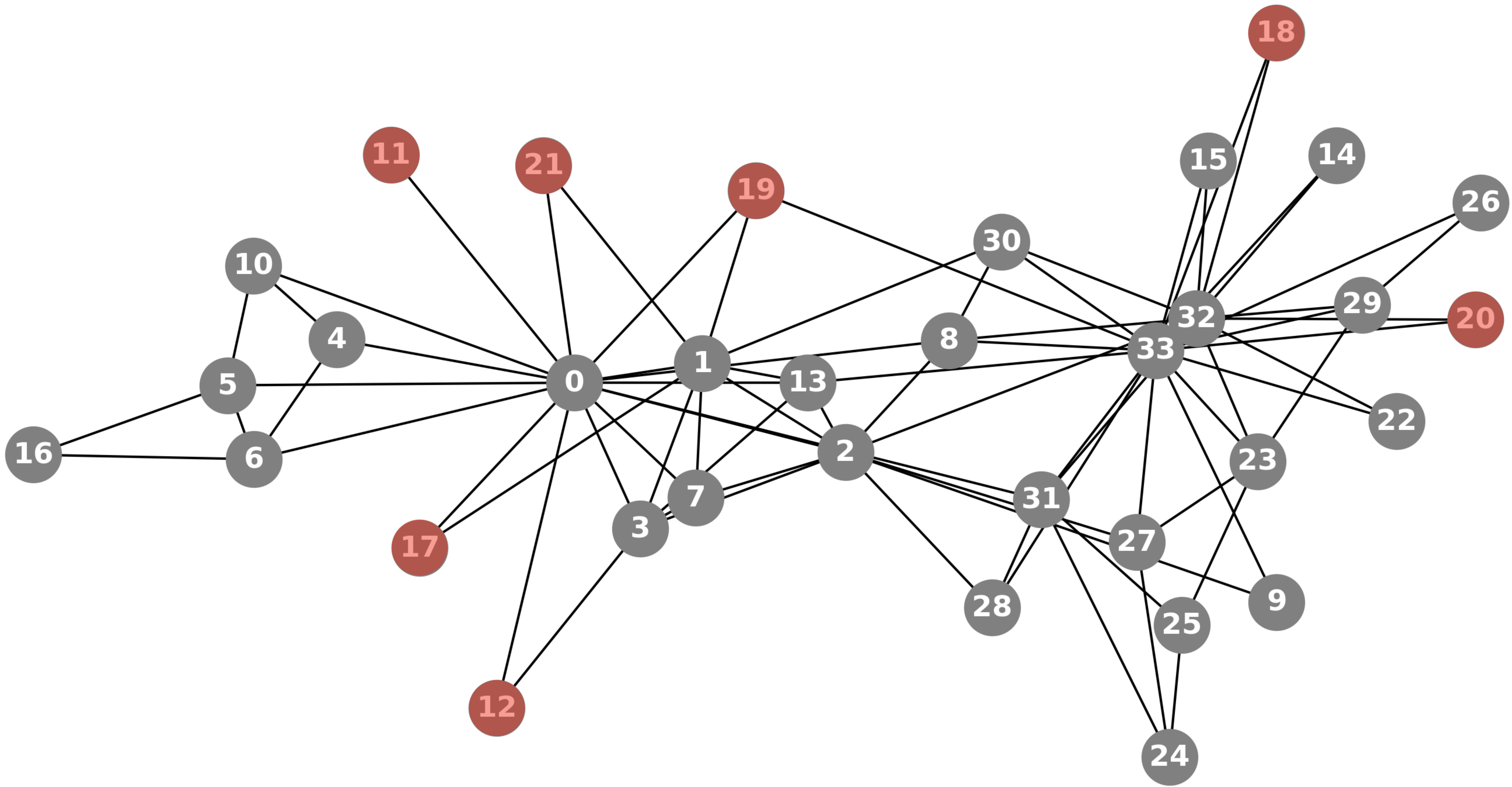}
    \label{fig:karate_network}}
    
    \subfloat[Data dependent radii (DDR) and model-dependent radii (MDR)]{
    \begin{tabular}{cc|cc}
Node & Degree & DDR & MDR \\
\hline
0 & 16  &  1.0 & 0.25\\ 
1 & 9 	& 0.41 & 0.13\\ 
2 & 10 	&  0.5 & 0.21\\ 
3 & 6 	& 0.45 & 0.55\\ 
4 & 3 	&  0.16 & 0.33\\ 
5 & 4 	&  0.22 & 0.42\\ 
6 & 4 	&  0.19 & 0.37\\ 
7 & 4 	&  0.24 & 0.1\\ 
8 & 5 	&  0.29 & 0.06\\ 
9 & 2 	&  0.0 & 0.0\\ 
10 & 3 	&  0.16 & 0.4\\ 
\textcolor{red}{\textbf{11}} & \textcolor{red}{\textbf{1}} & \textcolor{red}{\textbf{0.01}} & \textcolor{red}{\textbf{0.34}}\\ 
\textcolor{red}{\textbf{12}} & \textcolor{red}{\textbf{2}} & \textcolor{red}{\textbf{0.02}} & \textcolor{red}{\textbf{0.26}}\\ 
13 & 5 & 0.31 & 0.19\\ 
14 & 2 & 0.07 & 0.24\\ 
15 & 2 & 0.08 & 0.3\\ 
16 & 2 & 0.12 & 0.39\\ 
\textcolor{red}{\textbf{17}} & \textcolor{red}{\textbf{2}} & \textcolor{red}{\textbf{0.0}} & \textcolor{red}{\textbf{0.3}}\\ 
\textcolor{red}{\textbf{18}} & \textcolor{red}{\textbf{2}} & \textcolor{red}{\textbf{0.0}} & \textcolor{red}{\textbf{0.27}}\\ 
\textcolor{red}{\textbf{19}} & \textcolor{red}{\textbf{3}}& \textcolor{red}{\textbf{0.06}} & \textcolor{red}{\textbf{0.23}}\\ 
\textcolor{red}{\textbf{20}} & \textcolor{red}{\textbf{2}} & \textcolor{red}{\textbf{0.04}} & \textcolor{red}{\textbf{0.35}}\\ 
\textcolor{red}{\textbf{21}} & \textcolor{red}{\textbf{2}} & \textcolor{red}{\textbf{0.03}} & \textcolor{red}{\textbf{0.32}}\\ 
22 & 2 & 0.06 & 0.0\\ 
23 & 5 &  0.25 & 0.12\\ 
24 &	3	& 0.13 & 0.06\\ 
25 &	3	& 0.02 & 0.03\\ 
26 &	2	& 0.08 & 0.12\\ 
27 &	4	& 0.21 & 0.16\\ 
28 &	3	& 0.1 & 0.09\\ 
29 &	4	& 0.23 & 0.55\\ 
30 &	4	& 0.26 & 0.22\\ 
31 &	6	& 0.2 & 0.12\\ 
32 &	12	& 0.73 & 1.0\\ 
33 &	17	& 0.98 & 0.71\\ 
    \end{tabular}}
    \caption{Low degree nodes (\textcolor{red}{\textbf{bold}}) have low DDR possibly due to consistent edge reconstruction but high MDR indicating graph embedding functions may struggle to learn robust representations for low degree nodes.}
\end{figure}

To gain an intuitive understanding of what the radii represent, we analyze the well-known Karate Club network \cite{karate}. Figure \ref{fig:karate_network}(a) illustrates the network. The table in Figure \ref{fig:karate_network}(b) shows the data-dependent and model-dependent radii computed for each node in this network. Note that nodes 0 and 33 (representing the instructor and president, respectively) are key figures in their respective communities. In the case of data-dependent radii, most low-degree nodes (e.g., nodes 11, 12, 17, 18) exhibit small radii, indicating high confidence. This is likely because these nodes consistently recreate the same edges as the number of components increases during reconstruction. On the other hand, the model-dependent radii reveal another pattern, as this measure captures uncertainty in the learning algorithm during node classification. While nodes on the network's periphery, such as 11, 12, 17, and 18, display low data-dependent radii (i.e., the edges incident on these nodes are reliably reconstructed across graph views), they have large model-dependent radii. This suggests that graph embedding functions may struggle to learn robust representations for low-degree nodes. These radii are also visualized in Figure \ref{fig:karate} in the Appendix.

\section{Related Work}

Methods for handling uncertainty in graphs can be categorized into uncertain graphs, where edges carry probabilistic weights, or methods that explicitly capture data or model uncertainty. For uncertain graphs, Danesh et al.~\cite{danesh2022dgcu} use Gaussian embeddings with probabilistic proximity and similarity matrices for clustering. Hu et al.~\cite{hu2017embedding} use Jaccard similarity and random walks to embed uncertain graphs. Other methods include a k-nearest neighbor clustering \cite{potamias2010k}, a method to decompose the cores \cite{bonchi2014core}, and a method to compute shortest paths and centrality on uncertain graphs \cite{saha2021shortest}. 

For modeling and leveraging uncertainty in simple graphs, Vilnis and McCallum~\cite{vilnis2014word} use density-based distributed embeddings to learn representations in the space of Gaussian distributions similar to variational autoencoders. Vashishth et al.~\cite{vashishth2019confidence} uses covariance-based symmetric Mahalanobis distance between two nodes in the graph as label confidence scores during neighborhood aggregation in GCN. Wang et al.~\cite{wang2023rethinking} propose a recommendation system that estimates the randomness of missing data. Wang et al.~\cite{wang2023modeling} use epistemic and aleatoric uncertainties to boost exploration for recommender systems. Zhao et al.~\cite{zhao2020uncertainty} propose a graph-based kernel Dirichlet distribution estimation method that uses uncertainty from labels of training nodes to predict node-level Dirichlet distributions. Z.-Y.~Liu et al.~\cite{liu2020uncertainty} extend Graph Gaussian Processes \cite{ng2018bayesian} by incorporating uncertainty and label smoothing using symmetric Mahalanobis distances between nodes. Y.~Liu et al.~\cite{liu2022ud} estimate model uncertainty using Monte Carlo dropout variational inference (MCDVI) followed by retraining using pruning parameters for nodes with high certainty and retraining on nodes with low certainty. Trivedi et al.~\cite{trivediaccurate} use the concept of anchoring to estimate model uncertainty. They propose node feature and hidden layer readout anchoring methods that tackle the concept of anchoring in the graph domain for GNNs. Munikoti et al.~\cite{munikoti2023general} estimate data uncertainty in graphs using Assumed Density Filtering. Feng et al.~\cite{feng2021uag} capture both data and model uncertainty and use them in an attention mechanism. Data uncertainty is captured in terms of node diversity, and model uncertainty is captured using MCDVI. Huang et al.~\cite{huang2024uncertainty} extends conformal predictions to graph-based models for guaranteed uncertainty estimates. Uncertainty can also be addressed using fuzzy theory by considering weights of a GNN as asymmetric triangular fuzzy numbers \cite{lin2024fuzzy} or Gaussian membership functions to compute membership degrees for hidden layer representations \cite{zhang2023fuzzy} or by constructing a fuzzy graph by fusing multiple views of the input graph \cite{zhou2022fuzzy}. For an extensive discussion on uncertainty in machine learning and graph neural networks, we refer the reader to \cite{zhou2022survey} and \cite{wang2024uncertainty}.  

Graph adversarial robustness and defenses can be broadly categorized into either data pre-processing methods or model architectural changes aimed at enhancing the robustness of learning algorithms. For an overview, we refer the reader to the survey in \cite{sun2022adversarial}.

For data augmentation methods, Suresh et al.~\cite{suresh2021adversarial} and Jovanovic et al.~\cite{jovanovic2021towards} use Graph Contrastive Learning (GCL) to train on multiple views of a graph. Wang et al.~\cite{wang2019adversarial} combine GCL with an adversarial discriminator to train robust representations. Wu et al.~\cite{wu2019adversarial} and Li et al.~\cite{li2024exploiting} both use similar approaches to remove edges with low Jaccard similarity between node features, while Luo et al.~\cite{luo2021learning} learn a parametrized model to learn to drop edges. Wang et al.~\cite{wang2024graph} locate tractive nodes and reshape their local structure to improve their representation. Liu et al.~\cite{liugraph} extend the Graph Diffusion Convolution (GDC) model by incorporating min-max optimization to defend against adversarial attacks. Ye et al.~\cite{yericci} train GNNs using graphs resampled from the Ricci flow metric to get robust representations. 

To make learning algorithms more robust, Zhu et al.~\cite{zhu2019robust} adopt Gaussian distributions as hidden representations of nodes in graph convolution layers. Xu et al.~\cite{xu2019topology} uses adversarial training with a min-max optimization objective. Pan et al.~\cite{pan2019learning} use a variational graph autoencoder (VGAE) and a discriminator network to learn robust representations. Zhang et al.~\cite{zhang2020defensevgae} also use VGAE to reconstruct an attacked graph, followed by using a vanilla GCN on the reconstructed graph. He et al.~\cite{he2024enhancing} combine topology-driven label propagation, Bayesian label transitions, and link analysis via random walks to make GNNs robust to topological perturbations. In terms of certifiable robustness, Bojchevski et al.\cite{bojchevski2020efficient} propose an efficient method to extend robustness certificates to discrete graph data that can certify any GNN and handles perturbations to both graph structure as well as node attributes.

\section{Conclusion}
We introduced REGE, an approach for improving graph embedding functions by incorporating both data and model uncertainty. Through eigen-decomposition and the student-teacher model, REGE systematically measures and utilizes data and model uncertainty, improving the robustness of graph embedding functions against adversarial attacks. Our experimental results validate the efficacy of REGE, showing that it outperforms existing methods in defending against adversarial attacks by 1.5\% on average across four datasets.

\newpage

\bibliographystyle{IEEEtran}
\bibliography{bib}

\begin{thebibliography}{10}
\providecommand{\url}[1]{#1}
\csname url@samestyle\endcsname
\providecommand{\newblock}{\relax}
\providecommand{\bibinfo}[2]{#2}
\providecommand{\BIBentrySTDinterwordspacing}{\spaceskip=0pt\relax}
\providecommand{\BIBentryALTinterwordstretchfactor}{4}
\providecommand{\BIBentryALTinterwordspacing}{\spaceskip=\fontdimen2\font plus
\BIBentryALTinterwordstretchfactor\fontdimen3\font minus \fontdimen4\font\relax}
\providecommand{\BIBforeignlanguage}[2]{{%
\expandafter\ifx\csname l@#1\endcsname\relax
\typeout{** WARNING: IEEEtran.bst: No hyphenation pattern has been}%
\typeout{** loaded for the language `#1'. Using the pattern for}%
\typeout{** the default language instead.}%
\else
\language=\csname l@#1\endcsname
\fi
#2}}
\providecommand{\BIBdecl}{\relax}
\BIBdecl

\bibitem{wang2021review}
J.~Wang, S.~Zhang, Y.~Xiao, and R.~Song, ``A review on graph neural network methods in financial applications,'' \emph{arXiv preprint arXiv:2111.15367}, 2021.

\bibitem{chatterjee2023improving}
A.~Chatterjee, R.~Walters, Z.~Shafi, O.~S. Ahmed, M.~Sebek, D.~Gysi, R.~Yu, T.~Eliassi-Rad, A.-L. Barab{\'a}si, and G.~Menichetti, ``Improving the generalizability of protein-ligand binding predictions with ai-bind,'' \emph{Nature Communications}, vol.~14, no.~1, p. 1989, 2023.

\bibitem{zhang2024hyganno}
W.~Zhang, Y.~Cui, B.~Liu, M.~Loza, S.-J. Park, and K.~Nakai, ``Hyganno: hybrid graph neural network--based cell type annotation for single-cell atac sequencing data,'' \emph{Briefings in Bioinformatics}, vol.~25, no.~3, p. bbae152, 2024.

\bibitem{ying2018graph}
R.~Ying, R.~He, K.~Chen, P.~Eksombatchai, W.~L. Hamilton, and J.~Leskovec, ``Graph convolutional neural networks for web-scale recommender systems,'' in \emph{ACM SIGKDD}, 2018, pp. 974--983.

\bibitem{hinton2015distilling}
G.~Hinton, ``Distilling the knowledge in a neural network,'' \emph{arXiv preprint arXiv:1503.02531}, 2015.

\bibitem{zhanggraph}
S.~Zhang, Y.~Liu, Y.~Sun, and N.~Shah, ``Graph-less neural networks: Teaching old mlps new tricks via distillation,'' in \emph{ICLR}, 2022.

\bibitem{angelopoulos2021gentle}
A.~N. Angelopoulos and S.~Bates, ``A gentle introduction to conformal prediction and distribution-free uncertainty quantification,'' \emph{arXiv preprint arXiv:2107.07511}, 2021.

\bibitem{bengio2009curriculum}
Y.~Bengio, J.~Louradour, R.~Collobert, and J.~Weston, ``Curriculum learning,'' in \emph{ICML}, 2009, pp. 41--48.

\bibitem{xu2019topology}
K.~Xu, H.~Chen, S.~Liu, P.-Y. Chen, T.-W. Weng, M.~Hong, and X.~Lin, ``Topology attack and defense for graph neural networks: an optimization perspective,'' in \emph{IJCAI}, 2019, pp. 3961--3967.

\bibitem{zugner_adversarial_2019}
D.~Z{\"u}gner and S.~G{\"u}nnemann, ``Adversarial attacks on graph neural networks via meta learning,'' in \emph{ICLR}, 2019.

\bibitem{liu2022towards}
Z.~Liu, Y.~Luo, L.~Wu, Z.~Liu, and S.~Z. Li, ``Towards reasonable budget allocation in untargeted graph structure attacks via gradient debias,'' \emph{NeurIPS}, pp. 27\,966--27\,977, 2022.

\bibitem{young2020bayesian}
J.-G. Young, G.~T. Cantwell, and M.~Newman, ``Bayesian inference of network structure from unreliable data,'' \emph{Journal of Complex Networks}, vol.~8, 2020.

\bibitem{peixoto2024network}
T.~P. Peixoto, ``Network reconstruction via the minimum description length principle,'' \emph{arXiv preprint arXiv:2405.01015}, 2024.

\bibitem{barron1998minimum}
A.~Barron, J.~Rissanen, and B.~Yu, ``The minimum description length principle in coding and modeling,'' \emph{IEEE Transactions on Information Theory}, vol.~44, 1998.

\bibitem{sawlani2021fast}
S.~Sawlani, L.~Zhao, and L.~Akoglu, ``Fast attributed graph embedding via density of states,'' in \emph{ICDM}.\hskip 1em plus 0.5em minus 0.4em\relax IEEE, 2021, pp. 559--568.

\bibitem{cucuringu2011localization}
M.~Cucuringu and M.~W. Mahoney, ``Localization on low-order eigenvectors of data matrices,'' \emph{arXiv preprint arXiv:1109.1355}, 2011.

\bibitem{mitrovic2009spectral}
M.~Mitrovi{\'c} and B.~Tadi{\'c}, ``Spectral and dynamical properties in classes of sparse networks with mesoscopic inhomogeneities,'' \emph{Physical Review E—Statistical, Nonlinear, and Soft Matter Physics}, vol.~80, no.~2, p. 026123, 2009.

\bibitem{adamic2005political}
L.~A. Adamic and N.~Glance, ``The political blogosphere and the 2004 us election: divided they blog,'' in \emph{ACM LinkKDD}, 2005, pp. 36--43.

\bibitem{zhou2022survey}
X.~Zhou, H.~Liu, F.~Pourpanah, T.~Zeng, and X.~Wang, ``A survey on epistemic (model) uncertainty in supervised learning: Recent advances and applications,'' \emph{Neurocomputing}, vol. 489, pp. 449--465, 2022.

\bibitem{hullermeier2021aleatoric}
E.~H{\"u}llermeier and W.~Waegeman, ``Aleatoric and epistemic uncertainty in machine learning: An introduction to concepts and methods,'' \emph{Machine Learning}, vol. 110, no.~3, pp. 457--506, 2021.

\bibitem{kingma2013auto}
D.~P. Kingma, ``Auto-encoding variational bayes,'' \emph{arXiv preprint arXiv:1312.6114}, 2013.

\bibitem{mccallum2000automating}
A.~K. McCallum, K.~Nigam, J.~Rennie, and K.~Seymore, ``Automating the construction of internet portals with machine learning,'' \emph{Information Retrieval}, vol.~3, pp. 127--163, 2000.

\bibitem{giles1998citeseer}
C.~L. Giles, K.~D. Bollacker, and S.~Lawrence, ``Citeseer: An automatic citation indexing system,'' in \emph{ACM Conference on Digital Libraries}, 1998, pp. 89--98.

\bibitem{zhu2019robust}
D.~Zhu, Z.~Zhang, P.~Cui, and W.~Zhu, ``Robust graph convolutional networks against adversarial attacks,'' in \emph{ACM SIGKDD}, 2019, pp. 1399--1407.

\bibitem{entezari2020all}
N.~Entezari, S.~A. Al-Sayouri, A.~Darvishzadeh, and E.~E. Papalexakis, ``All you need is low (rank) defending against adversarial attacks on graphs,'' in \emph{ACM WSDM}, 2020, pp. 169--177.

\bibitem{jin2020graph}
W.~Jin, Y.~Ma, X.~Liu, X.~Tang, S.~Wang, and J.~Tang, ``Graph structure learning for robust graph neural networks,'' in \emph{ACM SIGKDD}, 2020, pp. 66--74.

\bibitem{zhang2020gnnguard}
X.~Zhang and M.~Zitnik, ``Gnnguard: Defending graph neural networks against adversarial attacks,'' \emph{NeurIPS}, pp. 9263--9275, 2020.

\bibitem{wang2024graph}
H.~Wang, C.~Zhou, X.~Chen, J.~Wu, S.~Pan, Z.~Li, J.~Wang, and S.~Y. Philip, ``Graph structure reshaping against adversarial attacks on graph neural networks,'' \emph{IEEE TKDE}, 2024.

\bibitem{liugraph}
S.~Liu, J.~Chen, T.~Fu, L.~Lin, M.~Zitnik, and D.~Wu, ``Graph adversarial diffusion convolution,'' in \emph{ICML}, 2024.

\bibitem{yericci}
Z.~Ye, T.~Ma, C.-C. Ni, K.~S. Liu, J.~Gao, and C.~Chen, ``Ricci-gnn: Defending against structural attacks through a geometric approach,'' 2020.

\bibitem{gasteiger2019diffusion}
J.~Gasteiger, S.~Wei{\ss}enberger, and S.~G{\"u}nnemann, ``Diffusion improves graph learning,'' \emph{NeurIPS}, vol.~32, 2019.

\bibitem{ollivier2009ricci}
Y.~Ollivier, ``Ricci curvature of markov chains on metric spaces,'' \emph{Journal of Functional Analysis}, vol. 256, pp. 810--864, 2009.

\bibitem{karate}
W.~W. Zachary, ``An information flow model for conflict and fission in small groups,'' \emph{Journal of Anthropological Research}, vol.~33, no.~4, pp. 452–--473, 1977.

\bibitem{danesh2022dgcu}
M.~Danesh, M.~Dorrigiv, and F.~Yaghmaee, ``Dgcu: A new deep directed method based on gaussian embedding for clustering uncertain graphs,'' \emph{Computers and Electrical Engineering}, vol. 101, p. 108066, 2022.

\bibitem{hu2017embedding}
J.~Hu, R.~Cheng, Z.~Huang, Y.~Fang, and S.~Luo, ``On embedding uncertain graphs,'' in \emph{ACM CIKM}, 2017, pp. 157--166.

\bibitem{potamias2010k}
M.~Potamias, F.~Bonchi, A.~Gionis, and G.~Kollios, ``K-nearest neighbors in uncertain graphs,'' \emph{Proceedings of the VLDB Endowment}, vol.~3, no. 1-2, pp. 997--1008, 2010.

\bibitem{bonchi2014core}
F.~Bonchi, F.~Gullo, A.~Kaltenbrunner, and Y.~Volkovich, ``Core decomposition of uncertain graphs,'' in \emph{ACM SIGKDD}, 2014, pp. 1316--1325.

\bibitem{saha2021shortest}
A.~Saha, R.~Brokkelkamp, Y.~Velaj, A.~Khan, and F.~Bonchi, ``Shortest paths and centrality in uncertain networks,'' \emph{Proceedings of the VLDB Endowment}, vol.~14, no.~7, pp. 1188--1201, 2021.

\bibitem{vilnis2014word}
L.~Vilnis and A.~McCallum, ``Word representations via gaussian embedding,'' \emph{arXiv preprint arXiv:1412.6623}, 2014.

\bibitem{vashishth2019confidence}
S.~Vashishth, P.~Yadav, M.~Bhandari, and P.~Talukdar, ``Confidence-based graph convolutional networks for semi-supervised learning,'' in \emph{AISTATS}, 2019, pp. 1792--1801.

\bibitem{wang2023rethinking}
C.~Wang, F.~Feng, Y.~Zhang, Q.~Wang, X.~Hu, and X.~He, ``Rethinking missing data: Aleatoric uncertainty-aware recommendation,'' \emph{IEEE Transactions on Big Data}, 2023.

\bibitem{wang2023modeling}
X.~Wang and S.~Kad{\i}o{\u{g}}lu, ``Modeling uncertainty to improve personalized recommendations via bayesian deep learning,'' \emph{IJDSA}, 2023.

\bibitem{zhao2020uncertainty}
X.~Zhao, F.~Chen, S.~Hu, and J.-H. Cho, ``Uncertainty aware semi-supervised learning on graph data,'' \emph{NeurIPS}, pp. 12\,827--12\,836, 2020.

\bibitem{liu2020uncertainty}
Z.-Y. Liu, S.-Y. Li, S.~Chen, Y.~Hu, and S.-J. Huang, ``Uncertainty aware graph gaussian process for semi-supervised learning,'' in \emph{AAAI}, 2020, pp. 4957--4964.

\bibitem{ng2018bayesian}
Y.~C. Ng, N.~Colombo, and R.~Silva, ``Bayesian semi-supervised learning with graph gaussian processes,'' \emph{NeurIPS}, vol.~31, 2018.

\bibitem{liu2022ud}
Y.~Liu, X.~Ao, F.~Feng, and Q.~He, ``Ud-gnn: Uncertainty-aware debiased training on semi-homophilous graphs,'' in \emph{ACM SIGKDD}, 2022, pp. 1131--1140.

\bibitem{trivediaccurate}
P.~Trivedi, M.~Heimann, R.~Anirudh, D.~Koutra, and J.~J. Thiagarajan, ``Accurate and scalable estimation of epistemic uncertainty for graph neural networks,'' in \emph{ICLR}, 2024.

\bibitem{munikoti2023general}
S.~Munikoti, D.~Agarwal, L.~Das, and B.~Natarajan, ``A general framework for quantifying aleatoric and epistemic uncertainty in graph neural networks,'' \emph{Neurocomputing}, vol. 521, pp. 1--10, 2023.

\bibitem{feng2021uag}
B.~Feng, Y.~Wang, and Y.~Ding, ``Uag: Uncertainty-aware attention graph neural network for defending adversarial attacks,'' in \emph{AAAI}, 2021, pp. 7404--7412.

\bibitem{huang2024uncertainty}
K.~Huang, Y.~Jin, E.~Candes, and J.~Leskovec, ``Uncertainty quantification over graph with conformalized graph neural networks,'' \emph{NeurIPS}, vol.~36, 2024.

\bibitem{lin2024fuzzy}
Y.-N. Lin, H.-C. Cai, C.-Y. Zhang, H.-Y. Yao, and C.~P. Chen, ``Fuzzy neural network for representation learning on uncertain graphs,'' \emph{IEEE Transactions on Fuzzy Systems}, 2024.

\bibitem{zhang2023fuzzy}
C.-Y. Zhang, Y.-N. Lin, C.~P. Chen, H.-Y. Yao, H.-C. Cai, and W.-P. Fang, ``Fuzzy representation learning on graph,'' \emph{IEEE Transactions on Fuzzy Systems}, vol.~31, no.~10, pp. 3358--3370, 2023.

\bibitem{zhou2022fuzzy}
J.~Zhou, Q.~Zhang, S.~Zeng, and B.~Zhang, ``Fuzzy graph subspace convolutional network,'' \emph{IEEE TNNLS}, 2022.

\bibitem{wang2024uncertainty}
F.~Wang, Y.~Liu, K.~Liu, Y.~Wang, S.~Medya, and P.~S. Yu, ``Uncertainty in graph neural networks: A survey,'' \emph{arXiv preprint arXiv:2403.07185}, 2024.

\bibitem{sun2022adversarial}
L.~Sun, Y.~Dou, C.~Yang, K.~Zhang, J.~Wang, S.~Y. Philip, L.~He, and B.~Li, ``Adversarial attack and defense on graph data: A survey,'' \emph{IEEE TKDE}, pp. 7693--7711, 2022.

\bibitem{suresh2021adversarial}
S.~Suresh, P.~Li, C.~Hao, and J.~Neville, ``Adversarial graph augmentation to improve graph contrastive learning,'' \emph{NeurIPS}, vol.~34, pp. 15\,920--15\,933, 2021.

\bibitem{jovanovic2021towards}
N.~Jovanovi{\'c}, Z.~Meng, L.~Faber, and R.~Wattenhofer, ``Towards robust graph contrastive learning,'' \emph{arXiv preprint arXiv:2102.13085}, 2021.

\bibitem{wang2019adversarial}
S.~Wang, Z.~Chen, J.~Ni, X.~Yu, Z.~Li, H.~Chen, and P.~S. Yu, ``Adversarial defense framework for graph neural network,'' \emph{arXiv preprint arXiv:1905.03679}, 2019.

\bibitem{wu2019adversarial}
H.~Wu, C.~Wang, Y.~Tyshetskiy, A.~Docherty, K.~Lu, and L.~Zhu, ``Adversarial examples for graph data: deep insights into attack and defense,'' in \emph{IJCAI}, 2019, pp. 4816--4823.

\bibitem{li2024exploiting}
D.~Li, H.~Wu, M.~Xie, X.~Wu, Z.~Wu, and W.~Zhang, ``Exploiting global graph homophily for generalized defense in graph neural networks,'' \emph{arXiv preprint arXiv:2406.03833}, 2024.

\bibitem{luo2021learning}
D.~Luo, W.~Cheng, W.~Yu, B.~Zong, J.~Ni, H.~Chen, and X.~Zhang, ``Learning to drop: Robust graph neural network via topological denoising,'' in \emph{ACM WSDM}, 2021, pp. 779--787.

\bibitem{pan2019learning}
S.~Pan, R.~Hu, S.-f. Fung, G.~Long, J.~Jiang, and C.~Zhang, ``Learning graph embedding with adversarial training methods,'' \emph{IEEE Transactions on Cybernetics}, pp. 2475--2487, 2019.

\bibitem{zhang2020defensevgae}
A.~Zhang and J.~Ma, ``Defensevgae: Defending against adversarial attacks on graph data via a variational graph autoencoder,'' \emph{arXiv preprint arXiv:2006.08900}, 2020.

\bibitem{he2024enhancing}
S.~He, J.~Zhuang, D.~Wang, L.~Peng, and J.~Song, ``Enhancing the resilience of graph neural networks to topological perturbations in sparse graphs,'' \emph{arXiv preprint arXiv:2406.03097}, 2024.

\bibitem{bojchevski2020efficient}
A.~Bojchevski, J.~Gasteiger, and S.~G{\"u}nnemann, ``Efficient robustness certificates for discrete data: Sparsity-aware randomized smoothing for graphs, images and more,'' in \emph{ICML}, 2020.

\end{thebibliography}

\appendix


\section{Additional Results}

\subsection{Results for Intermediate Perturbation Levels}
We present results for the CoraML \cite{mccallum2000automating} dataset in Table \ref{tab:cora_tab} as well as for additional perturbation levels of 3\% and 7\% in Tables \ref{tab:citeseer_tab_appendix}, \ref{tab:cora_tab_appendix}, \ref{tab:coraml_tab_appendix}, and \ref{tab:polblogs_tab_appendix} for the Citeseer, Cora, CoraML and PolBlogs datasets, respectively. Observe that REGE consistently outperforms the competing methods across these perturbation ranges.

\subsection{Ablation study on additional datasets}
We present results for REGE and the variant REGE-NCT (\textbf{N}o \textbf{C}urriculum \textbf{T}raining), which excludes curriculum training for the CoraML and PolBlogs dataset in Table \ref{tab:ablation_table_2}. Observe that REGE-NCT still demonstrates competitive performance, surpassing several state-of-the-art methods. However, REGE consistently outperforms REGE-NCT, highlighting the effectiveness of curriculum learning. The results presented are for 10\% perturbation levels.

\begin{table*}[]
    \centering
    \begin{tabular}{c|cc|cc|cc}
        Method          & MinMax (1\%)    &MinMax(10\%)     & Meta (1\%)      & Meta(10\%)      & GraD (1\%) & GraD (10\%) \\
        \hline
        GCN             &    $.819\pm.003$    &   $.637\pm.004$  &   $.831\pm.003$        &   $.723\pm.004$     &    $.825\pm.003$    &       $.809\pm.002$ \\
        Adv-Train             & 	$.782\pm.001$ 	&	$.782\pm.002$  &	 $.781\pm.002$ 	&	$.775\pm.001$  & $.771\pm.001$ 	&	$.763\pm.001$  \\

RGCN            &    $.85\pm.002$     &   $.775\pm.005$  &   $\textbf{.852}\pm\textbf{.001}$      &   $.755\pm.004$     &    $\textbf{.859}\pm\textbf{.002}$  &       $\underline{.826\pm.002}$ \\
GCN-SVD         &    $.811\pm.003$    &   $.805\pm.002$  &   $.808\pm.003$        &   $.801\pm.003$     &    $.785\pm.004$    &       $.782\pm.003$ \\
GNNGuard        &    $\textbf{.855}\pm\textbf{.002}$  &   $\underline{.826\pm.005}$  &   $\underline{.85\pm.003}$         &   $.811\pm.004$     &    $.854\pm.003$    &       $.823\pm.002$ \\
ProGNN          &    $.797\pm.005$    &   $.794\pm.005$  &   $.797\pm.005$        &   $.758\pm.031$     &    $.787\pm.01$     &       $.745\pm.048$ \\
GADC            &    $.795\pm.008$    &   $.783\pm.01$   &   $.797\pm.008$        &   $.794\pm.007$     &    $.78\pm.01$      &       $.805\pm.008$ \\
GraphReshape    &    $.821\pm.008$    &   $.823\pm.008$  &   $.83\pm.007$         &   $.817\pm.006$     &    $.834\pm.001$    &       $.815\pm.0$ \\
Ricci-GNN       &    $.84\pm.004$     &   $.808\pm.004$  &   $.828\pm.003$        &   $.811\pm.005$     &    $.841\pm.004$    &       $.822\pm.004$ \\
\hline
REGE (D)        &    $\underline{.853\pm.002}$    &   $\textbf{.829}\pm\textbf{.005}$    &   $\textbf{.852}\pm\textbf{.003}$      &   $\underline{.845\pm.002}$     &    $\underline{.858\pm.004}$    &       $\textbf{.836}\pm\textbf{.002}$ \\
REGE (M)        &    $.851\pm.003$    &   $.822\pm.005$  &   $\textbf{.852}\pm\textbf{.003}$      &   $\textbf{.846}\pm\textbf{.003}$   &    $.855\pm.003$    &       $\textbf{.836}\pm\textbf{.002}$ \\
    \end{tabular}
    \caption{Node classification accuracy on CoraML averaged over 10 runs, with standard deviations. REGE-D (data-dependent radii) and REGE-M (model-dependent rRadii) consistently outperform competing methods. Best results are in bold, with the second-best underlined. Comparisons include older methods such as RGCN and GCN-SVD, and newer methods such as GADC and GraphReshape.}
    \label{tab:coraml_tab}
\end{table*}

\begin{table*}[]
    \centering
    \begin{tabular}{c|cc|cc|cc}
      Method            & MinMax (3\%)       &  MinMax(7\%)          & Meta (3\%)        & Meta(7\%)         & GraD (3\%)           & GraD (7\%) \\
      \hline
GCN                 &   $.672\pm.003$  &  $.607\pm.006$  &   $.684\pm.003$     &  $.668\pm.004$  &   $.66\pm.005$     &   $.666\pm.003$ \\ 
Adv-Train             & 	$.660\pm.001$ 	&	$.657\pm.002$  &	 $.660\pm.002$ 	&	$.659\pm.001$  & $.578\pm.001$ 	&	$.58\pm.001$  \\
RGCN                &   $.696\pm.009$  &  $.664\pm.02$   &   $.72\pm.006$      &  $.701\pm.02$   &   $.666\pm.004$    &   $.661\pm.004$ \\ 
GCN-SVD             &   $.671\pm.006$  &  $.673\pm.007$  &   $.684\pm.006$     &  $.686\pm.005$  &   $.639\pm.009$    &   $.647\pm.006$ \\ 
GNNGuard            &   $.714\pm.006$  &  $.706\pm.005$  &   $.709\pm.013$     &  $.69\pm.015$   &   $.689\pm.006$    &   $.693\pm.005$ \\ 
ProGNN              &   $.714\pm.008$  &  $.714\pm.006$  &   $.721\pm.004$     &  $.716\pm.004$  &   $.687\pm.005$    &   $.681\pm.005$ \\ 
GADC                &   $.731\pm.008$  &  $.73\pm.008$   &   \underline{$.736\pm.007$}     &  $.733\pm.005$  &   \underline{$.696\pm.004$}    &   $.694\pm.003$ \\ 
GraphReshape        &   $.712\pm.005$  &  $.700\pm.006$    &   $.702\pm.003$     &  $.710\pm.004$   &   $.692\pm.002$    &   \underline{$.697\pm.008$} \\ 
Ricci-GNN           &   \underline{$.72\pm.004$}   &  $.71\pm.007$   &   $.72\pm.005$      &  $.721\pm.006$  &   $.667\pm.006$    &   $.664\pm.006$ \\ 
\hline
REGE (D)            &   $\textbf{.745}\pm\textbf{.002}$    &  \underline{$.742\pm.003$}  &   $\textbf{.75}\pm\textbf{.003}$        &  \underline{$.746\pm.003$}  &   $\textbf{.716}\pm\textbf{.003}$  &   $\textbf{.711}\pm\textbf{.006}$ \\ 
REGE (M)            &   $.718\pm.054$  &  $\textbf{.744}\pm\textbf{.001}$    &   $\textbf{.75}\pm\textbf{.003}$        &  $\textbf{.747}\pm\textbf{.002}$    &   $.694\pm.053$    &   $.694\pm.03$ \\ 

    \end{tabular}
    \caption{Node classification accuracy on Citeseer averaged over 10 runs, with standard deviations. REGE-D (data-dependent radii) and REGE-M (model-dependent radii) consistently outperform competing methods. Best results are in bold, with the second-best underlined. Comparisons include older methods such as RGCN and GCN-SVD, and newer methods such as GADC and GraphReshape.}
    \label{tab:citeseer_tab_appendix}
\end{table*}

\begin{table*}[]
    \centering
    \begin{tabular}{c|cc|cc|cc}
   Method          & MinMax (3\%)        &  MinMax(7\%)     & Meta (3\%)            & Meta(7\%)          & GraD (3\%)           & GraD (7\%) \\
\hline
GCN             &   $.783\pm.003$  &  $.721\pm.006$  &   $.768\pm.003$    &   $.727\pm.004$  & $.797\pm.003$ &    $.78\pm.003$ \\ 
Adv-Train             & 	$.758\pm.003$ 	&	$.755\pm.001$  &	 $.759\pm.001$ 	&	$.761\pm.004$  & $.755\pm.002$ 	&	$.766\pm.001$  \\
RGCN            &   $.813\pm.004$  &  $.785\pm.006$  &   $.794\pm.003$    &   $.758\pm.004$  & \underline{$.819\pm.003$} &    $.809\pm.004$ \\ 
GCN-SVD         &   $.718\pm.004$  &  $.71\pm.007$   &   $.714\pm.007$    &   $.707\pm.006$  & $.704\pm.006$ &    $.714\pm.004$ \\ 
GNNGuard        &   $.828\pm.004$  &  $.783\pm.007$  &   $.798\pm.005$    &   $.746\pm.01$   & $.807\pm.003$ &    $.814\pm.006$ \\ 
ProGNN          &   \underline{$.830\pm.005$}   &  $\textbf{.831}\pm\textbf{.003}$    &   $.83\pm.005$     &   $.812\pm.004$  & $.791\pm.003$ &    $.817\pm.004$ \\ 
GADC            &   $.784\pm.004$  &  $.783\pm.005$  &   $.787\pm.004$    &   $.781\pm.004$  & $.775\pm.006$ &    $.787\pm.008$ \\ 
GraphReshape    &   \underline{$.83\pm.002$}   &  $.817\pm.007$  &   $.825\pm.003$    &   $.814\pm.004$  & $.817\pm.003$ &    $.815\pm.004$ \\ 
Ricci-GNN       &   $.829\pm.005$  &  $.810\pm.005$   &   $.813\pm.006$    &   $.792\pm.004$  & $.818\pm.004$ &    $.807\pm.006$ \\ 
\hline
REGE (D)        &   $\textbf{.838}\pm\textbf{.003}$    &  $.82\pm.006$   &   \underline{$.848\pm.003$}    &   $\textbf{.845}\pm\textbf{.003}$    & $\textbf{.821}\pm\textbf{.004}$ &  \underline{$.825\pm.003$} \\ 
REGE (M)        &   $\textbf{.838}\pm\textbf{.003}$    &  \underline{$.822\pm.006$}  &   $\textbf{.85}\pm\textbf{.003}$   &   \underline{$.843\pm.005$}  & $\textbf{.821}\pm\textbf{.003}$ &  $\textbf{.827}\pm\textbf{.005}$ \\

    \end{tabular}
    \caption{Node classification accuracy on Cora averaged over 10 runs, with standard deviations. REGE-D (data-dependent radii) and REGE-M (model-dependent radii) consistently outperform competing methods. Best results are in bold, with the second-best underlined. Comparisons include older methods such as RGCN and GCN-SVD, and newer methods such as GADC and GraphReshape.}
    \label{tab:cora_tab_appendix}
\end{table*}

\begin{table*}[]
    \centering
    \begin{tabular}{c|cc|cc|cc}
Method          & MinMax (3\%)       &  MinMax(7\%)     & Meta (3\%)            & Meta(7\%)          & GraD (3\%)        & GraD (7\%) \\
\hline
GCN             &   $.795\pm.003$  &  $.714\pm.004$  & $.809\pm.002$      &   $.763\pm.004$  & $.818\pm.002$   &  $.809\pm.003$ \\ 
Adv-Train             & 	$.782\pm.003$ 	&	$.781\pm.004$  &	 $.780\pm.003$ 	&	$.776\pm.001$  & $.765\pm.003$ 	&	$.767\pm.002$  \\
RGCN            &   \underline{$.844\pm.002$}  &  $.807\pm.004$  & $.841\pm.003$      &   $.797\pm.003$  & $\textbf{.853}\pm\textbf{.003}$     &  $\textbf{.837}\pm\textbf{.004}$ \\ 
GCN-SVD         &   $.81\pm.003$   &  $.804\pm.003$  & $.805\pm.003$      &   $.795\pm.003$  & $.785\pm.002$   &  $.789\pm.003$ \\ 
GNNGuard        &   $\textbf{.852}\pm\textbf{.003}$    &  $\textbf{.831}\pm\textbf{.003}$    & $.848\pm.004$      &   $.834\pm.006$  & $.843\pm.003$   &  $.831\pm.003$ \\ 
ProGNN          &   $.799\pm.006$  &  $.795\pm.008$  & $.796\pm.006$      &   $.78\pm.008$   & $.774\pm.007$   &  $.762\pm.017$ \\ 
GADC            &   $.791\pm.009$  &  $.792\pm.008$  & $.793\pm.008$      &   $.796\pm.007$  & $.791\pm.004$   &  $.79\pm.005$ \\ 
GraphReshape    &   $.826\pm.006$  &  $.824\pm.005$  & $.828\pm.006$      &   $.822\pm.0$    & $.839\pm.006$   &  $.825\pm.003$ \\ 
Ricci-GNN       &   $.824\pm.005$  &  $.815\pm.005$  & $.824\pm.003$      &   $.807\pm.006$  & $.832\pm.004$   &  $.832\pm.004$ \\ 
\hline
REGE (D)        &   \underline{$.844\pm.003$}  &  \underline{$.829\pm.004$}  & \underline{$.852\pm.003$}      &   \underline{$.843\pm.004$}  & $.843\pm.002$   &  $\textbf{.837}\pm\textbf{.003}$ \\ 
REGE (M)        &   $.841\pm.002$  &  $.823\pm.01$   & $\textbf{.854}\pm\textbf{.003}$        &   $\textbf{.846}\pm\textbf{.004}$    & \underline{$.844\pm.002$}   &  \underline{$.833\pm.002$} \\ 
    \end{tabular}
    \caption{Node classification accuracy on CoraML averaged over 10 runs, with standard deviations. REGE-D (data-dependent radii) and REGE-M (model-dependent radii) consistently outperform competing methods. Best results are in bold, with the second-best underlined. Comparisons include older methods such as RGCN and GCN-SVD, and newer methods such as GADC and GraphReshape.}
    \label{tab:coraml_tab_appendix}
\end{table*}

\begin{table*}[]
    \centering
    \begin{tabular}{c|cc|cc|cc}

Method          & MinMax (3\%)          &   MinMax(7\%)     & Meta (3\%)             & Meta(7\%)            & GraD (3\%)            & GraD (7\%) \\
\hline
GCN             &   $.93\pm.001$      &   $.897\pm.002$  &   $.836\pm.002$     &  $.775\pm.003$     &    $.837\pm.002$    &   $.815\pm.002$ \\ 
Adv-Train             & 	$.899\pm.001$ 	&	$.871\pm.001$  &	 $.885\pm.001$ 	&	$.881\pm.001$  & $.808\pm.002$ 	&	$.796\pm.001$  \\
RGCN            &   $.924\pm.001$     &   $.885\pm.002$  &   $.805\pm.003$     &  $.745\pm.003$     &    $.84\pm.002$     &   $.818\pm.005$ \\ 
GCN-SVD         &   $.932\pm.004$     &   $.892\pm.005$  &   $.929\pm.006$     &  $.919\pm.003$     &    $.857\pm.002$    &   $.854\pm.003$ \\ 
GNNGuard        &   \underline{$.934\pm.004$ }    &   $.89\pm.002$   &   $.81\pm.003$      &  $.757\pm.008$     &    $.849\pm.002$    &   $.814\pm.003$ \\ 
ProGNN          &   $.924\pm.026$     &   $.889\pm.032$  &   $.924\pm.037$     &  $.868\pm.057$     &    $.858\pm.009$    &   $.866\pm.007$ \\ 
GADC            &   $.512\pm.008$     &   $.512\pm.008$  &   $.512\pm.008$     &  $.512\pm.008$     &    $.499\pm.007$    &   $.499\pm.017$ \\ 
GraphReshape    &   $.921\pm.006$     &   $.875\pm.008$  &   $.809\pm.005$     &  $.73\pm.001$      &    $.837\pm.004$    &   $.819\pm.005$ \\ 
Ricci-GNN       &   $.926\pm.003$     &   \underline{$.902\pm.003$}  &   $.942\pm.004$     &  $.942\pm.007$     &    \underline{$.879\pm.007$}    &   $.868\pm.01$ \\ 
\hline
REGE (D)        &   $\textbf{.939}\pm\textbf{.002}$   &   $\textbf{.907}\pm\textbf{.005}$    &   \underline{$.951\pm.008$}     &  $\textbf{.953}\pm\textbf{.005}$   &    \underline{$.879\pm.003$}    &   \underline{$.87\pm.005$} \\ 
REGE (M)        &   $.927\pm.009$     &   $.901\pm.009$  &   $\textbf{.953}\pm\textbf{.008}$   &  \underline{$.95\pm.007$}      &    $\textbf{.881}\pm\textbf{.002}$  &   $\textbf{.876}\pm\textbf{.003}$ \\ 

    \end{tabular}
    \caption{Node classification accuracy on PolBlogs averaged over 10 runs, with standard deviations. REGE-D (data-dependent radii) and REGE-M (model-dependent radii) consistently outperform competing methods. Best results are in bold, with the second-best underlined. Comparisons include older methods such as RGCN and GCN-SVD, and newer methods such as GADC and GraphReshape.}
    \label{tab:polblogs_tab_appendix}
\end{table*}

\begin{table*}[h]
    \centering
    \begin{tabular}{c|ccc|ccc}
         
             &                  &   CoraML          &                   &                    &   PolBlogs         &                 \\
Method       &   MinMax         &   Meta-Attack     &   GraD            &   MinMax           &   Meta-Attack      &   GraD          \\
\hline
REGE-NCT (D) & $.749\pm.021$    &   $.737\pm.006$    &   $.817\pm.003$   &   $.874\pm.002$    &    $.721\pm.006$   &  $.826\pm.004$  \\
REGE (D)     & $\textbf{.829}\pm\textbf{.005}$    &   $\textbf{.845}\pm\textbf{.002}$    &   $\textbf{.836}\pm\textbf{.002}$   &   $\textbf{.89}\pm\textbf{.004}$    &    $\textbf{.95}\pm\textbf{.005}$   &  $\textbf{.865}\pm\textbf{.003}$  \\
\hline
REGE-NCT (M) & $.811\pm.002$    &   $.776\pm.005$   &   $.817\pm.002$   &   $.878\pm.002$    &    $.732\pm.003$   &  $.824\pm.002$   \\
REGE (M)     & $\textbf{.822}\pm\textbf{.005}$     &   $\textbf{.846}\pm\textbf{.003}$   &   $\textbf{.836}\pm\textbf{.002}$   &   $\textbf{.88}\pm\textbf{.006}$    &    $\textbf{.942}\pm\textbf{.017}$   &   $\textbf{.861}\pm\textbf{.004}$ \\
    \end{tabular}
    \caption{Node classification results from the ablation study for CoraML and PolBlogs networks. The REGE-NCT variant, which excludes curriculum learning, still demonstrates superior performance compared to other state-of-the-art methods as shown in Tables \ref{tab:polblogs_tab} and \ref{tab:coraml_tab}. However, REGE consistently outperforms REGE-NCT, highlighting the effectiveness of incorporating both radii and curriculum learning into the model. The results presented are for 10\% perturbation levels averaged across 10 runs.}
    \label{tab:ablation_table_2}
\end{table*}

\begin{figure*}[h]
    \centering
    \includegraphics[width=1\linewidth]{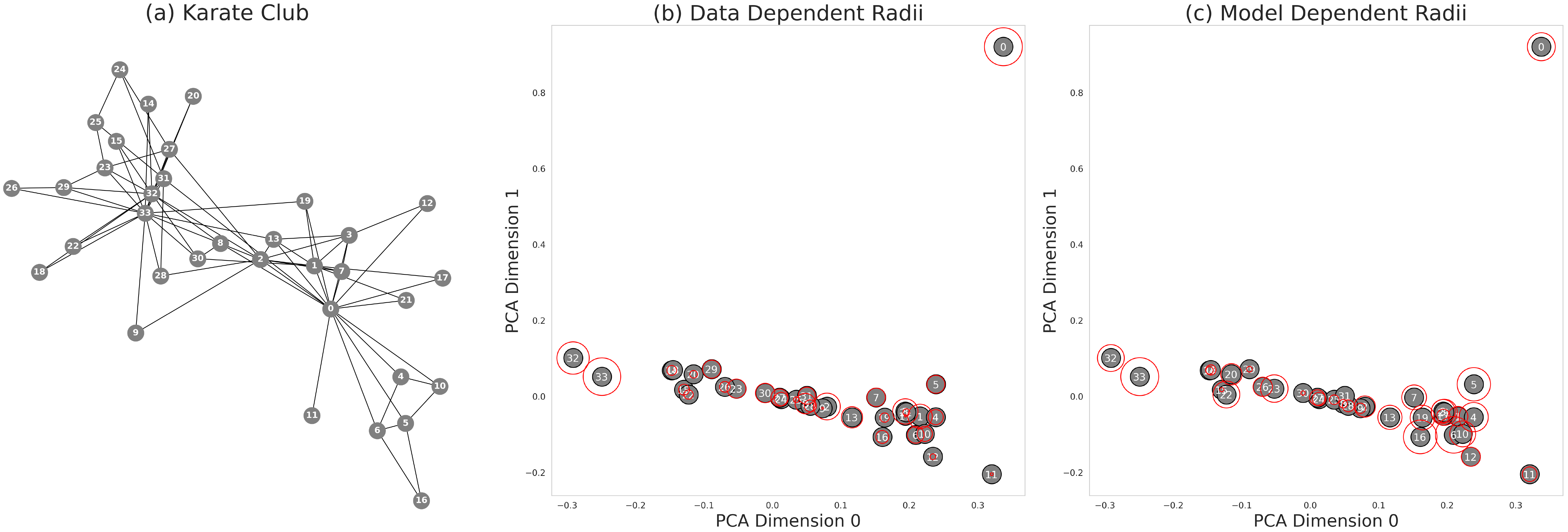}
    \caption{(a) Visualization of the Karate Club network with (b) data-dependent and (c) model-dependent radii. In (b), low-degree nodes (e.g., 11, 12, 17, 18) exhibit small radii, indicating high confidence due to consistent edge reconstruction. In (c), nodes on the network's periphery show larger model-dependent radii, reflecting uncertainty in the learning algorithm, especially for low-degree nodes.}
    \label{fig:karate}
\end{figure*}

\subsection{Karate Club Visualization}

To gain a visual understanding of what the radii represent, we embed the karate club network into 16 dimensions using a GCN and visualize it in 2 dimensions using PCA in Figures \ref{fig:karate}(b) and (c), where each node is surrounded by the data-dependent radii depicted in red in \ref{fig:karate}(b) and model-dependent radii in \ref{fig:karate}(c). Nodes 0 and 33 (representing the instructor and president, respectively) are key figures in their respective communities. In the case of data-dependent radii (\ref{fig:karate}(b)), most low-degree nodes (e.g., nodes 11, 12, 17, 18) exhibit small radii, indicating high confidence. This is likely because these nodes consistently recreate the same edges as the number of components increases during reconstruction. On the other hand, model-dependent radii (\ref{fig:karate}(c)) reveal another pattern, as this measure captures uncertainty in the learning algorithm during node classification. For instance, nodes on the network's periphery, such as 11, 12, 17, and 18, display low data-dependent radii (i.e., the edges incident on these nodes are reliably reconstructed across graph views) but large model-dependent radii, suggesting that graph embedding functions may struggle to learn robust representations for low-degree nodes.

\subsection{Varying Components}
\begin{figure*}[h]
    \centering
    \includegraphics[width=1\linewidth]{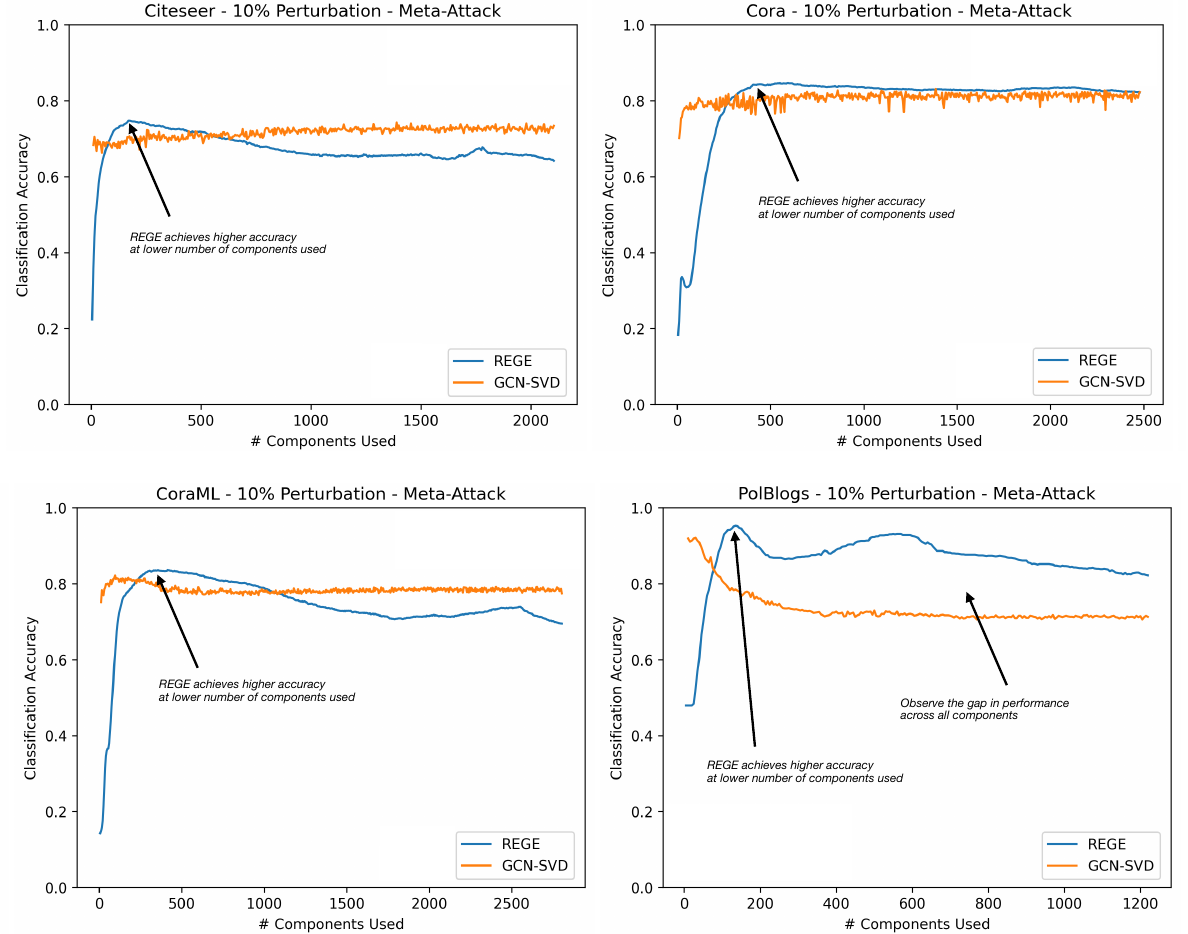}
    \caption{Comparison between REGE and GCN-SVD under varying numbers of components for datasets perturbed by Meta Attack at a 10\% perturbation rate. REGE consistently demonstrates higher accuracy across all datasets with fewer components.}
    \label{fig:svd_compare}
\end{figure*}

We conduct a detailed comparison between REGE and GCN-SVD. We select GCN-SVD because it can also vary the number of components used during training. Figure \ref{fig:svd_compare} illustrates the results, with the $x$-axis representing the number of components used to reconstruct the graph and the $y$-axis showing node classification accuracy. Across various datasets, REGE consistently achieves higher accuracy with fewer components. 

\subsection{Hyperparameters}
For the reconstruction of each $G_i$ from its eigen decomposition, we scale the reconstructed adjacency matrix $A_i$ to values between 0 and 1 with a threshold of $0.5$ to discretize the matrix. 
When generating graph views, we increment a number of components by 5 as we observe no performance gain from finer increments. Concretely, given the initial number of components $q$, $G_1$ is reconstructed using $q$ components, $G_2$ is reconstructed using $q + 5$ components, and so on. 

The GCNs have 16 hidden dimensions and 2 GCN layers with a dropout of $0.5$ after each layer. In the student-teacher framework, the teacher model uses the same GCN described above, and the student model is a simple MLP with 3 hidden layers and 1024 neurons in each layer with a dropout of $0.5$ after the first two layers. 

For training REGE, we train on each graph $G_i$ for 100 epochs and use an early stopping criteria with the patience set to 25 graph views. Our implementation is available at \url{https://github.com/zohairshafi/REGE}.

\end{document}